\documentclass[10pt,twocolumn,letterpaper]{article}

\usepackage[pagenumbers]{wacv} 

\usepackage{graphicx}
\usepackage{amsmath}
\usepackage{amssymb}
\usepackage{booktabs}
\usepackage[accsupp]{axessibility}  

\usepackage[pagebackref,breaklinks,colorlinks]{hyperref}

\usepackage[capitalize]{cleveref}
\crefname{section}{Sec.}{Secs.}
\Crefname{section}{Section}{Sections}
\Crefname{table}{Table}{Tables}
\crefname{table}{Tab.}{Tabs.}

\def\wacvPaperID{} 
\def\confName{WACV}
\def\confYear{2024}

\begin{document}

\title{FacadeNet: Conditional Facade Synthesis via Selective Editing}

\author{Yiangos Georgiou$^1$ \quad Marios Loizou$^1$ \quad Tom Kelly$^2$ \quad Melinos Averkiou$^1$ \\
$^1$Univesity of Cyprus/CYENS CoE, Cyprus {\tt\small \{ygeorg01, mloizo11, maverk01\}@ucy.ac.cy} \\
$^2$KAUST, Saudi Arabia  {\tt\small thomas.kelly@kaust.edu.sa}
}
\twocolumn[{%
 \renewcommand\twocolumn[1][]{#1}%
 \maketitle
 \thispagestyle{empty}
 \vspace{-6mm}
 \centering
 \includegraphics[width=\textwidth]{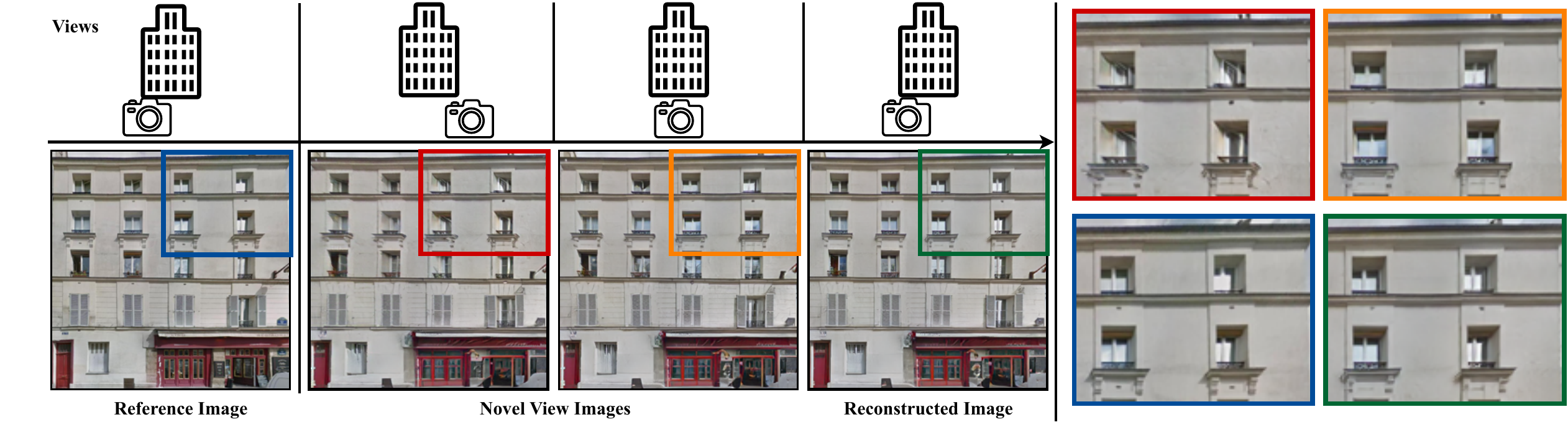}
 \vspace{-8mm}
 \captionof{figure}
{\textit{Left:} Utilizing a reference facade image (bottom row, left column) and relative camera 
position information (top row), our method generates novel facades from varied viewpoints, all 
while preserving the reference image's structure and style (centered columns). Additionally, 
our approach faithfully reconstructs the reference facade (right column). \textit{Right:} Zoomed 
facade regions highlight our method's capacity to modify critical facade elements, like windows, 
across diverse viewpoints (red and orange regions). Furthermore, our approach accurately 
reconstructs (green region) the reference facade (blue region).}
\label{fig:teaser}
\vspace{2mm}
}]

\maketitle

\begin{abstract}
\vspace{-3mm}
We introduce FacadeNet, a deep learning approach for synthesizing building facade images from diverse viewpoints.
Our method employs a conditional GAN, taking a single view of a facade along with the desired viewpoint information
and generates an image of the facade from the distinct viewpoint. To precisely modify \textit{view-dependent} 
elements like windows and doors while preserving the structure of \textit{view-independent} components such as 
walls, we introduce a \textit{selective editing module}. This module leverages image embeddings extracted from a 
pretrained vision transformer. Our experiments demonstrated state-of-the-art performance on building facade 
generation, surpassing alternative methods.
\end{abstract}  

\vspace{-5mm}
\section{Introduction}
In urban planning, architectural design, and historical preservation, there is an increasing demand for rich visual representations of building facades, as it facilitates more comprehensive visual analyses, interactive 3D visualizations such as virtual tours, and digital archiving of structures~\cite{debevec1996modeling,remondino2006image}. A conventional approach to capturing building facades involves taking photographs from different viewpoints. However, this approach is often constrained by practical limitations, such as the unavailability of multiple vantage points, especially in densely built urban environments. Moreover, obtaining a large dataset of images from varying views is time-consuming and expensive. In light of these challenges, synthesizing novel views of building facades from a single image emerges as a compelling alternative.

We tackle the problem of synthesizing images of a building's facade from novel viewpoints, given a single image of the facade taken from an arbitrary viewpoint. Synthesizing novel views of a facade given a single image entails estimating the appearance of the facade as seen from viewpoints different from the original image. This problem has been a subject of interest in computer graphics and computational photography due to its wide range of applications and inherent challenges associated with geometric and photometric consistency~\cite{seitz1996view,zhou2016stereo}

Traditional methods have relied on 3D reconstruction techniques~\cite{hartley2003multiple}, which involve extensive manual intervention and are not easily scalable. Learning-based approaches, especially conditional Generative Adversarial Networks (cGANs), have recently been explored for view synthesis~\cite{flynn2016deepstereo,isola2017imagetoimage}. However, state-of-the-art methods such as StyleGAN2~\cite{Karras2020ada} and Swapping Autoencoder~\cite{park2020swapping} that rely on style-content separation, often fail to decouple view information from structural properties of the facade. 

In contrast, our approach, \textit{FacadeNet}, addresses these challenges through a novel \textit{selective editing module}, which enables finer control over the generation process. It guides the generation by computing a \textit{selective editing mask}, allowing the alteration of view-dependent elements (e.g. windows) while keeping view-independent elements (e.g. walls) intact.
FacadeNet takes as input a single image of a building facade together with the desired view information in the form of a view tensor, and uses a conditional GAN equipped with our novel selective editing module to synthesize an image of the facade from a different viewpoint. 

Computing a selective editing mask could be simplified, if a semantic segmentation of the input facade image is available, as in SPADE~\cite{park2019SPADE} and SEAN~\cite{zhu2020sean}.
Unfortunately such semantic segmentation masks are not available for the vast majority of facade images captured in the wild. Using a pretrained network to generate such semantic masks is possible, but these will always be imprecise, especially at boundaries. Inspired by recent advances in the explainability of large pretrained vision transformer models ~\cite{amir2021deep}, we hypothesize that a \textit{selective editing mask} could be computed by combining, in a learnable network module, features from deep layers of such models. Our novel \textit{selective editing module} (Section~\ref{sec:methodology}) takes as input image features obtained by a pretrained DINO model~\cite{caron2021emerging}, and learns an optimal weighting of them in order to compute a selective editing mask. This mask then drives the synthesis of a facade image where view dependent parts such as windows and doors are edited according to the provided view information, while view independent parts such as the walls remain fixed. 

Through a series of experiments (Section~\ref{sec:experiments}), we demonstrate that our method outperforms quantitatively and qualitatively competing works, as well as strong baselines such as having access to ground truth semantic segmentation masks, on the large LSAA facade image dataset~\cite{lsaa}. Further ablations motivate the design choices for our selective editing module as well as some technical implementation details.

In summary, the main contributions of this paper are:
\vspace{-2mm}
\begin{enumerate}
    \item Introducing a novel selective editing module within a conditional GAN, that enables real-time synthesis of novel facade views from a single arbitrary image.\vspace{-2mm}
    \item Demonstrating through rigorous evaluation that FacadeNet outperforms state-of-the-art alternatives in single-image facade view synthesis.\vspace{-2mm}
    \item Present a comprehensive ablation study, unveiling the importance of different components in the FacadeNet architecture.\vspace{-2mm}
    \item Showcasing two applications of our novel approach in (i) eliminating rectification artifacts in facade images extracted from panoramic street-view images, and (ii) real-time texturing of simple 3D building models with dynamic camera-dependent facade views.
\end{enumerate}

\section{Related Work}

Our work lies within the broader realm of view synthesis, with a particular focus on synthesizing novel views of building facades via conditional GANs. In this section, we briefly review key areas of related work, including facade image analysis, traditional view synthesis, learning-based view synthesis, and conditional GANs for image generation.

\vspace*{-3mm}
\paragraph*{Facade Image Analysis.} Building facades have been a subject of interest due to their role in urban planning, architectural design, and 3D modeling. Debevec et al.~\cite{debevec1996modeling} used facade images for architectural scene modeling, while
Remondino and El-Hakim~\cite{remondino2006image} emphasized image-based 3D modeling for archival and historical preservation, focusing on facades. Datasets like eTrims~\cite{etrims}, CMP Facade dataset~\cite{cmpfacade}, and Graz50~\cite{graz50}, have been developed to facilitate research on facade analysis and modeling, aiding tasks such as facade segmentation, object detection, and 3D reconstruction~\cite{teboul2010segmentation}. However, these datasets predate recent deep learning advancements and are relatively small, thus are unsuitable for our purposes. The recent LSAA dataset of Zhu et al~\cite{lsaa} provides a large set of rectified facade images with various metadata including annotated view information and is thus used to train and evaluate our network.

\vspace*{-3mm}
\paragraph*{Traditional View Synthesis} View synthesis involves generating new scene images from viewpoints different from the available ones. Classic methods include view morphing by Seitz and Dyer~\cite{seitz1996view}, a technique that generates intermediate views of a scene by blending and warping two or more images, and Debevec et al.~\cite{debevec1996modeling}, who focused on creating photorealistic models of architectural scenes from images. Traditional approaches focused on geometric and photometric consistency to synthesize novel views. Hartley and Zisserman~\cite{hartley2003multiple} provide an extensive exploration of the geometry involved in multiple view synthesis. McMillan and Bishop~\cite{mcmillan1995plenoptic} proposed the image-based rendering technique, which involved blending of different views. While these methods were ground-breaking, they often require extensive manual effort and are not easily scalable. In contrast, our method is fully-automated, scalable, works in real-time, and does not rely on any hand-engineered features or correspondences to generate novel views of a facade.

\vspace*{-3mm}
\paragraph*{Learning-based View Synthesis} The advent of deep learning led to learning-based methods gaining popularity for view synthesis tasks. Hedman and Kopf~\cite{hedman2018deep} used a deep neural network to synthesize motion blur and refocus images. Flynn et al.~\cite{flynn2016deepstereo} introduced DeepStereo, which predicts new views from large natural imagery datasets. Zhou et al.~\cite{zhou2016stereo} advanced this domain with a multiplane image representation for stereo magnification. Neural Radiance Fields~\cite{mildenhall2020nerf, martinbrualla2020nerfw,Barron2021MipNeRFAM} have recently taken the view synthesis and reconstruction research areas by storm. However, they necessitate training on multiple images per facade, requiring training from scratch for each new facade, and often lack real-time speed, although recent methods offer vast improvements~\cite{hedman2021snerg}. Diffusion models~\cite{diffusion,ddpm,nichol2021improved,dhariwal2021diffusion} are a valid recent alternative to GANs, but they are slower and harder to control. Diffusion-based generative models \cite{ho2020denoising, song2020improved} have gained substantial attention for surpassing GANs in FID scores, particularly in unconditionally generated tasks like ImageNet \cite{ho2022cascaded} and super-resolution \cite{saharia2021image}. However, image editing poses a more intricate challenge for diffusion models. Recent advancements in both conditional \cite{saharia2021palette} and unconditional \cite{choi2021ilvr} diffusion models have tackled this, yielding high-quality results. Here, we undertake a comparative analysis between the performance of Palette\cite{saharia2021palette} and FacadeNet. In contrast, our method offers real-time performance, generalizes to unseen facades, and only requires one facade image as input. 

\vspace*{-3mm}
\paragraph*{Conditional Generative Adversarial Networks} GANs have been used for various image generation tasks, and conditional GANs, in particular, have been successful in image-to-image translation tasks. The Pix2Pix network by Isola et al.~\cite{isola2017imagetoimage} is a well-known example of using conditional GANs for image translation. 
StyleGAN~\cite{karras2019style} introduced a novel approach that utilizes a learned constant feature map and a generated latent code $z$ to control the output image features.
StyleGAN2~\cite{Karras2020ada} further enhances this concept with AdaIN~\cite{huang2017arbitrary} layer, which adjusts image channels to unit variance and zero mean, retaining channel statistics. It then incorporates style through scaling and shifting, guided by conditional information. Karras et al~\cite{karras2020analyzing} introduced an alternative approach that retains scale-specific control, eliminating undesired artifacts while preserving result quality.
Image2StyleGAN~\cite{Abdal2019Im2St, Abdal2020Im2St2} aimed to overcome a disadvantage of StyleGAN-based approaches to manipulate reference images. However, these approaches are slow due to a pre-processing step that iteratively predicts the latent code to reconstruct the reference image.

Another line of work to disentangle an image's latent space was introduced in the Swapping Autoencoder paper~\cite{park2020swapping} where two discriminators are used to disentangle style and structure from a reference input image. IcGAN~\cite{Perarnau2016} on the other hand used two inverse encoders to allow changing of the conditional and latent vectors separately. StarGAN~\cite{choi2018stargan} is one of the most common multi-domain image-to-image translation method which tries to improve the scalability of GAN models by using a single generator to generate all available domains. To extend StarGAN~\cite{choi2018stargan} to a multi-modal approach, StarGANv2~\cite{choi2020stargan} replaced the domain label with a domain specific style latent code that can represent diverse styles for each of the available domains. The SPADE network~\cite{park2019SPADE} introduced a novel conditional batch normalization specifically modified for images. An improved approach that enables multi-modal controllable style change for each different input was described in the SEAN network~\cite{zhu2020sean}.
Our method is most closely related to StyleGAN2~\cite{Karras2020ada} and Swapping Autoencoder~\cite{park2020swapping}, that rely on style-content decoupling. Encoding facade view information as \textit{style} is not enough however, since conditional GANs often fail to decouple view information from structural properties of the facade. As a result, they tend to hallucinate elements (e.g. windows or balconies), or add noise and other artifacts. FacadeNet tackles the shortcomings of these methods through a novel \textit{selective editing module}, which computes a \textit{selective editing mask} based on pretrained DINO~\cite{caron2021emerging} features to guide the generation, allowing it to focus on altering only the view-dependent elements (e.g., doors, balconies, windows).

Currently, 3D-aware synthesis techniques \cite{chan2022efficient, gu2021stylenerf, zhou2021cips} exhibit training efficiency and sampling capabilities comparable to 2D Generative Adversarial Networks (GANs). However, their effectiveness relies on meticulously curated datasets with aligned structures and scales, such as those for human or cat faces. A recent study \cite{skorokhodov20233d}, circumvented previous challenges, notably dependence on known camera poses. This workaround trains models to learn pose distributions from single-view data, rather than relying on multi-view observations. Similarly, our proposed approach eliminates the need for known camera poses and can be trained on sparse single-view image data.

\begin{figure*}[t]
  \centering
  \includegraphics[scale=0.5]{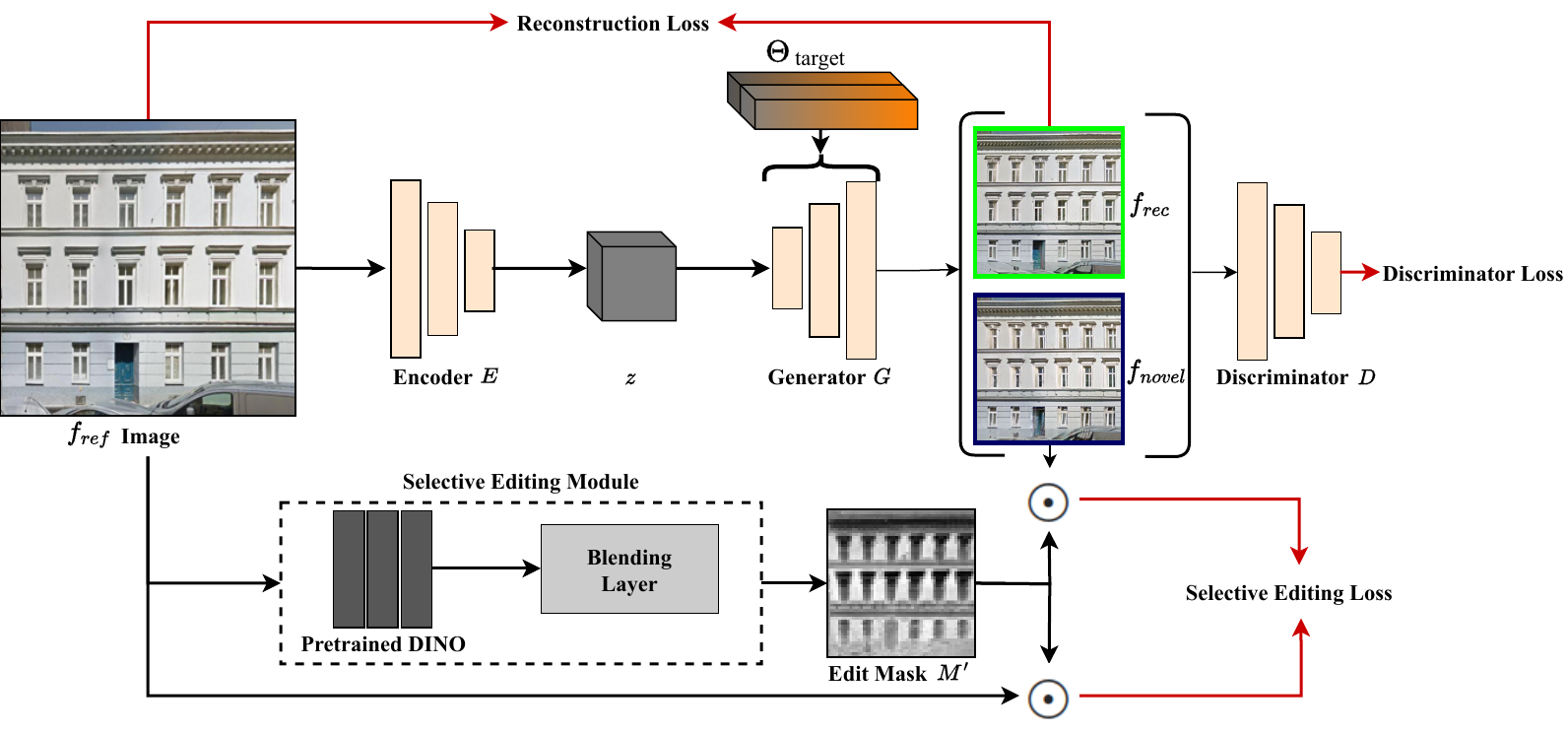}
  \vspace*{-4mm}
  \caption{Our training procedure creates latent tensors $z = E(f_{ref})$ which capture the reference facade's style and structure. These $z$ tensors form the basis for diverse view generation from a single reference image $f_{ref}$ with conditional information $\theta$. The generator utilizes the latent code $z$ and the target view direction $\theta$ to produce new samples $G(z, \theta_{target})$. Our approach accurately reconstructs $f_{ref}$ by aligning $\theta_{target}$ with the same viewpoint of $f_{ref}$, while varying the $\theta_{target}$ it generates novel views $f_{novel}$.  To ensure high-quality and consistent facade reconstruction from various viewpoints, our model employs multiple loss functions. These encompass 
 $L_1$ reconstruction and conditional discriminator losses, enhancing fidelity and structural awareness. Furthermore, we introduce a selective editing module that employs prior information (DINO ViT features) to extract a selective editing mask. This mask designates editable components during novel view synthesis, contributing to the precision of the process.}
  \label{fig:model}
  \vspace*{-5.5mm}
\end{figure*}

\section{Method}\label{sec:methodology}

Our approach focuses on synthesizing building facades with varying viewing angles, based on a reference facade image 
$\mathbf{f}_{ref} \sim \mathbf{F} \subset \mathbb{R}^{HxWx3}$ and horizontal and vertical angle vectors $\mathbf{\theta}_h 
\sim \Theta_H \in [-1, +1]^W$ and $\mathbf{\theta}_v \sim \Theta_V \in [-1, +1]^H$. By conditioning the generation process 
on the new view direction controlled by the input angle vectors, our method, called \textit{FacadeNet}, produces facades 
that exhibit realistic modifications in semantic components like windows and doors. To ensure authenticity, we incorporate a 
discriminator $D$ that enforces plausible changes in these components. Preserving the overall facade structure is crucial. 
Thus, our network is designed to faithfully reconstruct areas such as walls that remain unchanged regardless of the viewing 
direction. To achieve this, we utilize the feature embeddings of a self-supervised vision transformer \cite{caron2021emerging} 
to construct a semantic-aware mask, which guides the reconstruction process. Moreover, we employ a reconstruction loss to 
enhance the structural accuracy of the generated facades. In the following sections, we first provide an overview of our 
proposed architecture in Section \ref{subsec:facadenet}. We then delve into how we enforce the generation of structurally 
aware and novel building facades in Section \ref{subsec:structural awareness} and Section \ref{subsec:novel_reconstruction}, 
respectively. Finally, in Section \ref{subsec:implementation_details}, we present the implementation details of our approach.

\subsection{FacadeNet architecture}
\label{subsec:facadenet}
To generate building facades with specified viewing directions based on a reference image $\mathbf{f}_{ref}$ and
guided by the target viewing vector $\mathbf{\theta}_{target} = [\mathbf{\theta}_h, \mathbf{\theta}_v]$, we employ a 
\textit{task-specific conditional} GAN architecture \cite{isola2017imagetoimage}. As 
illustrated in Figure \ref{fig:model}, FacadeNet comprises an autoencoder network. First, the encoder $E$, inspired by 
\cite{park2020swapping}, takes the reference building facade as input and produces a latent tensor $\mathbf{z}$ 
with spatial dimensions, encoding both the structural and texture information present in the image $\mathbf{f}_{ref}$:
\vspace{-2mm}
\begin{equation}
    \mathbf{z} = E(\mathbf{f}_{ref})
\vspace{-2mm}
\end{equation}
Next, the conditional generator $G$, following \cite{karras2020analyzing}, utilizes the latent tensor 
$\mathbf{z}$ to synthesize a facade that aligns with the desired viewing direction, using the target viewing vector 
$\mathbf{\theta}_{target}$:
\vspace{-2mm}
\begin{equation}
    \mathbf{f}_{novel} = G\big(E(\mathbf{f}_{ref}), [\mathbf{\theta}_h, \mathbf{\theta}_v]\big)
\vspace{-1mm}
\end{equation}
The generated facade $\mathbf{f}_{novel}$ is crafted by combining the encoded reference image embedding $\mathbf{z}$
and the target viewing information. To ensure realistic results, the discriminator $D$ assesses the novel facade.
The target viewing vector is also incorporated into the discriminator to enforce that the generated facade aligns with
the desired viewing direction, by adopting the conditional discriminator idea of \cite{isola2017imagetoimage}.

\vspace*{-2mm}
\subsection{Structural awareness}
\label{subsec:structural awareness}
In our specific task, we aim to modify the high-frequency areas within the facade image, as they have a greater 
impact on influencing the viewing direction of the building being represented. At the same time, we want to preserve 
the low-frequency structural components typically found in building facades, such as flat surfaces like walls. These 
components remain visually consistent regardless of the observer's viewing position.
\vspace{-3mm}
\paragraph*{Accurate reconstruction.} In line with the principles of the classic autoencoder \cite{dimmethod}, we aim to acquire a mapping between the latent code $\mathbf{z} \sim \mathbf{Z}$ and the image $\mathbf{f}_{ref} 
\sim \mathbf{F} \subset \mathbb{R}^{HxWx3}$. To this end, we employ an image reconstruction loss that compares the input 
facade $\mathbf{f}_{ref}$ with the reconstructed facade $\mathbf{f}_{rec}$ conditioned on the \textit{ground truth} target
viewing vector $\mathbf{\theta}_{target}^{gt}$, which corresponds to the direction of the camera that originally captured 
the input image:
\begin{equation}
    \small
    \mathcal{L}_{rec}(E, G; \mathbf{\theta}_{t}^{gt}) = \mathbb{E}_{\mathbf{f}_{ref} \sim \mathbf{F}}\big[|\mathbf{f}_{ref} - G\big(E(\mathbf{f}), [\mathbf{\theta}_h^{gt}, \mathbf{\theta}_v^{gt}]\big)|\big]
    \label{eq:rec_loss}
\end{equation}
During the optimization process of the autoencoder, this loss facilitates the acquisition of accurate and informative latent 
representations for building facades.

\begin{figure*}[tbp]
  \centering
  \includegraphics[width=\linewidth]{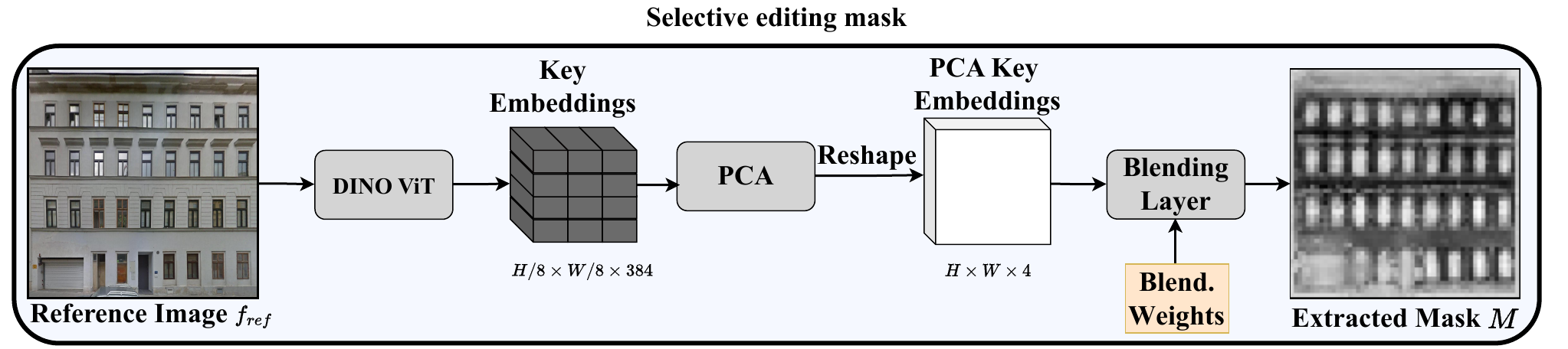}
\vspace*{-7mm}
  \caption{This figure depicts the step-by-step procedure for calculating selective editing masks for a given image $f_{ref}$. We commence by employing a pre-trained DINO ViT model to extract localized key embeddings that encapsulate crucial visual features within the image. Once DINO key embeddings are obtained, we apply PCA (Principal Component Analysis) to retain high-frequency information from these embeddings. Lastly, our adaptive blending layer utilizes the PCA embeddings, enabling optimal feature combination extraction. This process yields a single-channel selective mask denoted as $\pmb{M}$.}
  \label{fig:mask}
  \vspace*{-4mm}
\end{figure*}

\vspace*{-2mm}
\paragraph*{Selective Editing Module.} Our primary goal is to selectively modify areas within the input facade that have a 
significant imp\-act on the building's viewing direction while disregarding regions that maintain visual consistency 
relative to the capturing camera's orientation. To accomplish this, we introduce the \textit{Selective Editing Module} that 
constructs the \textit{selective editing mask}, effectively isolating these areas within the facade image. Drawing inspiration from the 
findings of \cite{amir2021deep}, we initially pass the input facade through a 
pre-trained self-supervised vision transformer \cite{caron2021emerging} and extract the key representations of the last attention 
layer. To retain high-frequency information from the underlying data distribution, we employ principal component analysis 
(PCA) on these embeddings By selecting the top four principal components $\mathbf{V}_{pc} = [\mathbf{v}_1, \mathbf{v}_2, 
\mathbf{v}_3, \mathbf{v}_4]$, we capture the most salient features. To create the selective 
editing mask, we blend these 
selected principal components using a linear combination approach. More specifically, we employ a set of learnable blending 
weights $\mathbf{W} \in \mathbb{R}^4$ and apply the sigmoid function $\sigma(\cdot)$ to obtain the final mask:
\vspace{-1mm}
\begin{equation}
    \mathbf{M} = \mathbf{V}_{pc} \cdot \sigma(\mathbf{W})
    \label{eq:mask}
\vspace{-1mm}
\end{equation}
This formulation enables the network to learn the optimal linear combination of the principal components, effectively 
highlighting the most prominent elements within the facade, such as windows and doors (see Figure \ref{fig:mask}). The extracted selective editing mask is semantic-aware, since it captures semantic areas in the facade image that are view-dependent.

In addition to the reconstruction loss (Eq \ref{eq:rec_loss}), which penalizes modifications across the entire reconstructed 
image, we introduce the \textit{Selective Editing Loss} that capitalizes on the selective editing mask $\mathbf{M}$. This loss 
compares the masked input facade $\mathbf{f}_{ref}$ with the masked novel facade $\mathbf{f}_{novel}$ whose viewing 
direction is altered according to the conditional target viewing direction $\mathbf{\theta}_{target}^{novel}$, rather than 
the ground truth direction of the reference facade. While the extracted mask $\mathbf{M}$ highlights editable areas within 
the facade, the primary objective of this loss is to retain the appearance of view-independent components. To achieve this, 
we utilize the complement of the mask, denoted as $\mathbf{M}' = 1 - \mathbf{M}$. By employing $\mathbf{M}'$, we effectively 
ignore high-frequency areas, enabling the network to preserve the viewing direction of the view-independent structural 
elements during the synthesis of novel facades with varying viewing directions:
\begin{equation}
    \small
    l_{edit}(E, G; \mathbf{\theta}_{t}^{n}) = || \mathbf{f}_{ref} \odot \mathbf{M}' - G\big(E(\mathbf{f}_{ref}, \mathbf{\theta}_{t}^{n})\big) \odot \mathbf{M}' ||
\end{equation}
Furthermore, as part of our training process, we synthesize $n$ novel facades for each input image and apply the selective 
editing loss. This step is crucial in ensuring that the appearance of view-independent components remains consistent across 
various novel facades with different viewing directions:
\vspace{-3mm}
\begin{equation}
    \mathcal{L}_{edit}(E, G; [\mathbf{\theta}_{t}^{n_1}, \mathbf{\theta}_{t}^{n_2}, \cdots, \mathbf{\theta}_{t}^{n_k}]) = \frac{1}{k} \sum_{i=1}^k l_{edit}^{(i)}
\end{equation}

\vspace*{-4mm}
\subsection{Novel reconstruction}
\label{subsec:novel_reconstruction}
When it comes to image editing, a crucial aspect is to modify the latent representation of an input image in a way 
that ensures the novel reconstruction appears both authentic and aligns with the provided conditional information.
Simultaneously, this representation should be able to faithfully and easily reconstruct the input image. To address these 
requirements, we introduce two key loss components: the \textit{View-dependent loss} and the \textit{View-consistent 
loss}. These losses play a vital role in guiding the network to generate realistic novel facades while preserving the 
fidelity of the reference facade and its original viewing direction, enforced by the discriminator $D$.

\vspace*{-2mm}
\paragraph*{View-dependent loss.} This loss facilitates the network's ability to synthesize novel facades that appear 
authentic and visually consistent with the specified viewing direction. By incorporating the conditional information, the 
network learns to modify the relevant components of the facade, ensuring the alterations align with the intended changes in 
the viewing perspective. Following the synthesis process during of the selective editing loss, we utilize a conditional 
adversarial loss that guides the generation of multiple novel facades based on the conditional target viewing 
directions $\mathbf{\theta}_{target}^{novel_i}$:
\begin{equation}
    \small
    \mathcal{L}_{GANd}^{dep}(E, G, D; \mathbf{\theta}_{t}^{n_i}) = \mathbb{E}_{\mathbf{f}_{ref} \sim \mathbf{F}, \mathbf{\theta} \sim \Theta} [-log(D(\mathbf{f}_{n}^i, \mathbf{\theta}))]
\end{equation}

\vspace{-4mm}
\paragraph*{View-consistent loss.} In parallel, the View-consistent loss serves to maintain the faithfulness of the 
reconstruction process for the reference facade. By minimizing the impact of viewing direction changes on the entire image, 
we ensure the reconstruction remains as close as possible to the original facade, thereby preserving its original viewing 
direction. The non-saturating adversarial loss \cite{goodfellow2014generative} for the generator $G$ and the encoder $E$ is 
computed as:
\vspace{-2mm}
\begin{equation}
    \mathcal{L}_{GAN}^{cons}(E, G, D; \mathbf{\theta}_{t}^{gt}) = \mathbb{E}_{\mathbf{f}_{ref} \sim \mathbf{F}}[-log(D(\mathbf{f}_{rec}, \mathbf{\theta}_{t}^{gt}))]
\vspace{-2mm}
\end{equation}

\subsection{Implementation Details} 
\label{subsec:implementation_details}

In this section we provide the implementation details regarding our training process. For the construction process of the target viewing vectors please see our supplementary material.

To optimize our network, we combine the structural awareness and novel reconstruction losses using linear weights::
\begin{equation}
    \mathcal{L}_{total} = \lambda_1 \mathcal{L}_{rec} + \lambda_2 \mathcal{L}_{edit} + \lambda_3 \mathcal{L}_{GAN}^{dep} + \lambda_4 \mathcal{L}_{GAN}^{cons}
\end{equation}
We empirically set the weights as $\lambda_1=\lambda_2=3$ and $\lambda_3=\lambda_4=0.5$, effectively balancing the 
importance of each loss component. For training, we employ the Adam optimizer with a learning rate of $0.001$. Our network 
is trained on four NVIDIA V100 GPUs using a batch size of $32$ images at a resolution of $256\times256$. The training 
process consists of 15 million iterations, taking approximately 4 days to complete. Remarkably, our trained model achieves 
an impressive average generation speed of 62 frames per second (fps) for a batch size of 32 images. We also refer readers to our project page with source code for more details. 
\footnote{\emph{\href{https://ygeorg01.github.io/FacadeNet}{ygeorg01.github.io/FacadeNet} includes our code and trained models}.}


\begin{table*}[t]
\centering
\begin{tabular}{||c|c c c c c c ||} 
 \hline
 Method & LPIPS-alex$\downarrow$  & LPIPS-vgg$\downarrow$ & FID$_{rec}\downarrow$ & FID$_{novel}\downarrow$ & PSNR$\uparrow$ & SSIM$\uparrow$ \\ [0.5ex] 
 \hline\hline
  StyleGAN2-ADA \cite{Karras2020ada}  & \textbf{\textendash} & \textbf{\textendash} & 22.52 & \textbf{\textendash} & 20.591 & 0.467 \\
 Palette \cite{saharia2022palette}  & 0.259 & 0.401 & 23.367 & 22.829 & 17.881 & 0.332 \\
 3DGP  \cite{skorokhodov20233d} & 0.186 & 0.347 & 35.918 & 33.005 & 14.673 & 0.201 \\
 Swapping-AE \cite{park2020swapping}  & 0.198 & 0.386 & 10.55 & 15.64 & 22.24 & 0.668 \\
 $FacadeNet_{base}$ & 0.174 & 0.296 & 10.59 & 9.91 & \textbf{24.13} & 0.69 \\
 $FacadeNet_{full}$ & \textbf{0.119} & \textbf{0.240} & \textbf{9.601} & \textbf{8.327} & 23.866 & \textbf{0.714} \\
 \hline
\end{tabular}
\vspace*{-1mm}
\caption{This table presents a comprehensive comparison between our model baseline $FacadeNet_{base}$, StyleGAN2\cite{karras2020analyzing}, Palette \cite{saharia2022palette}, 3DGP  \cite{skorokhodov20233d}, swapping-autoencoder\cite{park2020swapping} and $FacadeNet$. The results clearly demonstrate the superiority of our task-specific model across various evaluation criteria, including reconstruction quality, novel view synthesis quality, and consistency. To assess the reconstruction image quality, we employ $FID_{rec}$, $PSNR$, and $SSIM$ metrics. Regarding novel view image quality we rely on $FID_{novel}$, while we measure the inter-view consistency with $LPIPS-\{alex,vgg\}$ metrics. Our final model $FacadeNet$ outperforms previous approaches by a significant margin.}
\vspace*{-4mm}
\label{table:sota_comp}
\end{table*}

\section{Evaluation} \label{sec:experiments}

This section presents a comprehensive evaluation of FacadeNet's performance, combining qualitative and quantitative analysis. We begin by discussing the dataset used for our experiments in Section \ref{subsec:dataset}, providing important context for the subsequent evaluations. In Section \ref{subsec:quant}, we conduct a comparative analysis of our approach against state-of-the-art GAN-based and diffusion-based models. To assess performance, we employ various metrics, illustrating the superiority of FacadeNet.
To conclude our evaluation, Section \ref{subsec:applications} presents two novel applications specifically tailored to urban environments. These applications serve as powerful demonstrations of the versatility and potential impact of FacadeNet. Through these innovative use cases, we highlight the practical value and wider implications of our research. Please see our supplementary material for an in-depth ablation study of our design choices, where we emphasize the effectiveness of our selective editing module. Moreover, we provide examples of facade interpolation under varying viewing directions.

\vspace*{-2mm}
\subsection{Dataset} \label{subsec:dataset}
We make use of a rectified facade dataset sourced from \cite{lsaa} to obtain genuine rectified facade images. This dataset comprises approximately $23,000$ images. 
The dataset  provides various attributes from four distinct categories: \textit{Metadata} (including geographic properties like longitude, latitude, city, and building information), \textit{homography} (encompassing view angle and homography error), \textit{semantic attributes} (such as windows, balcony, and door area), and \textit{semantic embedding} (capturing similarities between samples). These attributes offer valuable insights into various scenarios. 

In our particular case, we leverage the homography attributes, particularly the view angle and the size of the cropped facade, to generate pairs of facade images and view direction targets (see supplementary). These targets represent the viewpoint for each given facade image, thereby enabling accurate processing of images regarding the angle information.

\vspace*{-1mm}
\subsection{Comparisons with other methods} \label{subsec:quant}
In this section, we perform a comparison between our approach and the state-of-the-art approaches related to our work. We use styleGAN2-ADA\cite{Karras2020ada}, Palette\cite{saharia2021palette}, 3DGP\cite{skorokhodov20233d} and Swapping-autoencoder\cite{park2020swapping} to extract metrics that measure the quality and consistency of facade reconstruction and novel view synthesis.
To assess the performance of facade reconstruction, we utilize several metrics including $PSNR$, $SSIM$, and $FID_{rec}$. These metrics allow us to evaluate both the image quality and the structural similarity between the reference and generated reconstructed images. Additionally, for evaluating the quality of novel view synthesis, we rely on the $FID_{novel}$ score. In terms of consistency and perceptual similarity, we employ $LPIPS_{vgg}$ and $LPIPS_{alex}$ metrics\cite{Karras2019stylegan2}, which aim to measure the smoothness of viewpoints interpolation. A low $LPIPS$ score indicates that the image patches are perceptually similar which implies fewer alternations in structure and style between viewpoints.

To measure StyleGAN2\cite{Karras2020ada} and 3DGP\cite{skorokhodov20233d} reconstruction quality, we project the reference image to the latent space in an iterative manner. We set the maximum number of optimization iterations to $1000$ and $4000$ respectively. We further modified the Swapping-Encoder\cite{park2020swapping} model to disentangle structure and novel direction, thereby enabling novel view manipulation and synthesis, and trained Palette\cite{saharia2021palette} given as conditional input the angle-view maps and the reference image to follow our task's approach. Further, we introduce two versions of our network: $FacadeNet_{base}$ serves as the baseline, which relies solely on L1 and adversarial losses to optimize its performance. On the other hand, $FacadeNet_{full}$ represents the enhanced version of $FacadeNet$, incorporating all the design choices discussed in section \ref{sec:methodology}, including the utilization of the selective editing module.

To assess the capabilities of StyleGAN2\cite{Karras2020ada} and 3DGP reconstruction \cite{skorokhodov20233d}, we employ an iterative approach to project the reference image into the latent space. We set the maximum optimization iterations to $1000$ for StyleGAN2 and $4000$ for $3DGP$. Furthermore, we have customized the Swapping-Encoder model\cite{park2020swapping} to disentangle structure and novel direction, thereby enabling novel view manipulation and synthesis. Additionally, we have trained Palette\cite{saharia2021palette} using angle-view maps and the reference image as conditional inputs, aligning with our task's approach. Moreover, we introduce two versions of our network $FacadeNet_{base}$ serves as the baseline, relying solely on L1 and adversarial losses to optimize its performance. Conversely, $FacadeNet_{full}$ represents the enhanced iteration of FacadeNet, incorporating all the design choices discussed in Section \ref{sec:methodology}, including the utilization of the selective editing module.

Table \ref{table:sota_comp} showcases a comprehensive comparison of various models. Among them, $FacadeNet_{base}$ and $FacadeNet_{full}$ achieves higher reconstruction quality, outperforming alternative approaches with $PSNR$ scores of $24.13$ and $23.86$, and $SSIM$ values of $0.69$ and $0.714$, respectively. Notably, the performance of $Swapping-AE$\cite{park2020swapping} and $Palette$\cite{saharia2021palette} surpasses that of $3DGP$\cite{skorokhodov20233d}.

In terms of $FID$ scores, our $FacadeNet$ approach generates more realistic samples than other models. Particularly in novel view facade synthesis, the $FID$ score difference increases noticeably. Our design choices effectively elevate facade synthesis, resulting in robust artifact-free novel views. $FacadeNet_{full}$ outperforms swapping-AE\cite{park2020swapping} by 1.04 $FID$ score points for $FID_{rec}$, and this gap widens to 7.32 for novel view synthesis $FID_{novel}$.

A central advancement of $FacadeNet$ centers around the coherence among diverse viewpoints. To quantify this aspect, we employ the $LPIPS$ metric. Notably, compared to other methods, $3DGP$\cite{skorokhodov20233d} excels in consistency, by achieving the highest scores. Building upon this, $FacadeNet_{full}$ further surpasses $3DGP$\cite{skorokhodov20233d} by improving $LPIPS_{alex}$ and $LPIPS_{vgg}$ scores by $0.05$ and $0.88$ respectively.

\begin{figure*}[tbp]
  \centering
  \includegraphics[width=.99\linewidth, height=10cm]{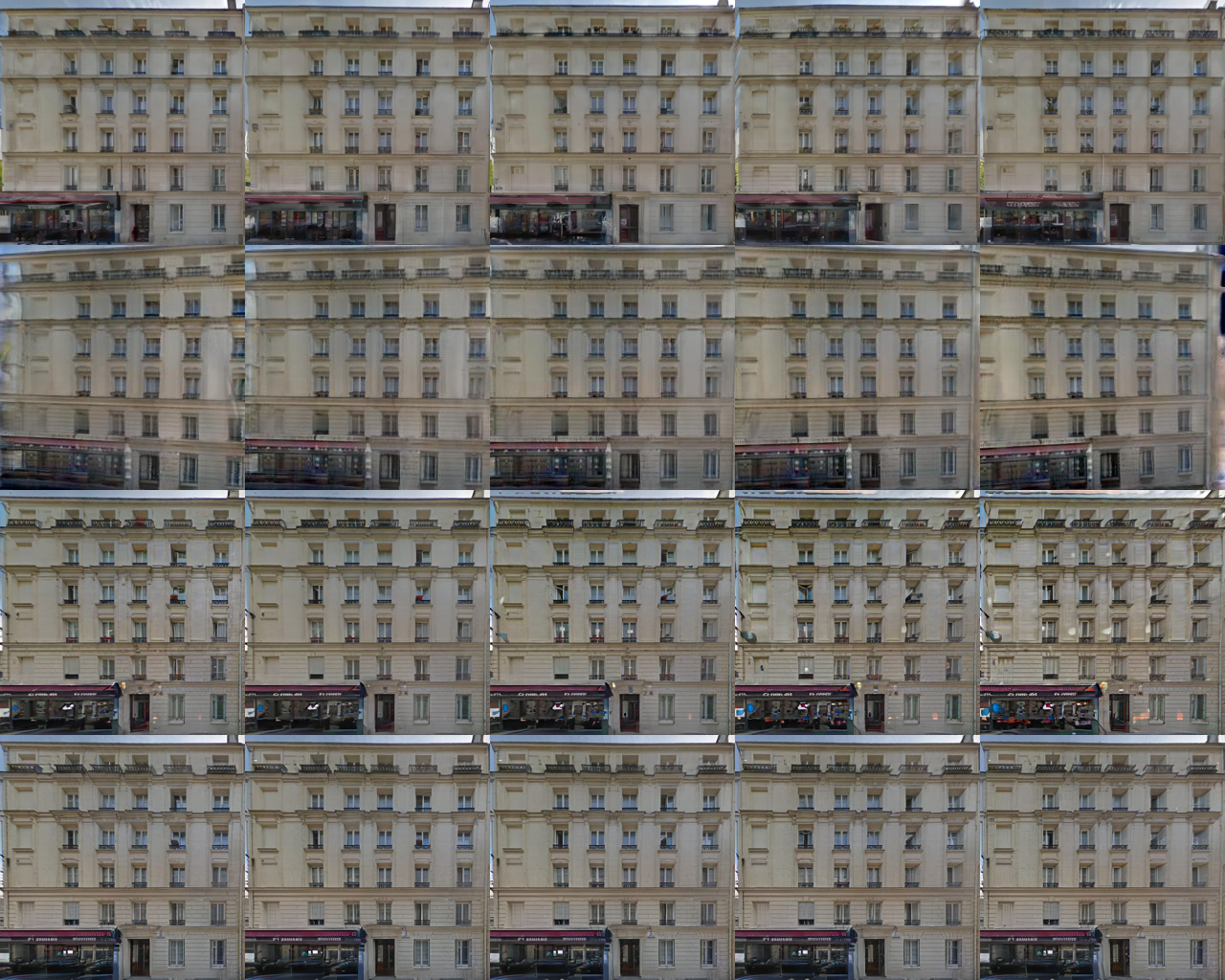}
\vspace*{-2mm}
  \caption{This figure presents qualitative comparisons between \textit{Palette}~\cite{saharia2021palette} ($1^{st}$ row), \textit{3DGP}~\cite{skorokhodov20233d} ($2^{nd}$ row), \textit{swapping-AE}~\cite{park2020swapping} ($3^{rd}$ row) and $FacadeNet_{full}$ ($4^{th}$ row). \textit{Palette} and \textit{3DGP} are unable to generate fine details as the generation is combined with novel view synthesis. Notably, artifacts become apparent in the output generated by the $swapping-AE$ model across varying viewing angles. In contrast, $FacadeNet_{full}$'s results  demonstrate a higher level of robustness, effectively preserving the structural details. More results are displayed in the supplementary.}
  \label{fig:interpolation}
\vspace*{-5mm}
\end{figure*}

\subsection{FacadeNet Applications}\label{subsec:applications}

\paragraph*{Problematic Rectified Facade Improvement} Our main assumption to improve problematic facade images, is that 0-view angle difference facades are closer to ortho-rectified images that contain minimum distortion or other artifacts. We define as problematic the group of facades that their mean view-angle value is higher than $60^{\circ}$. Those facades have extreme orientations either to the left or right and contain large areas of missing information on doors and windows or other assets.

Moreover, our methodology manages to generate hidden information that is not visible in the reference facades(top images) due to the originally captured view angle. Our generator $G$ manages to generate new information for the unseen parts of the input facade images with identical style and structure which is essential for the consistency between input and output facade images. Examples are provided in the supplementary.

\vspace{-3mm}
\paragraph*{Real-Time Textures for Urban Scenes} Following the facade view interpolation experiments we developed an application where we represent a 3D city environment by using interactive textures created by our generative model. More specifically we create a large scene that contains simple cubes $b \sim B$ that represent the buildings and we use reference facades $f \sim F$ as textures. As we navigate around the city the view direction targets between the camera and the points are computed. The computed view angle maps are used as conditional view direction targets $\theta_{target} = \{\theta_{h}, \theta_{v}\}$ to alter the orientation of each texture map accordingly. The equations to compute the view direction target in 3D environment are as follows:
\vspace{-2mm}
\begin{align}
d &= 
\begin{array}{ll}
    p - c
\end{array} \\
\theta_h^f(d, n)&= 
\begin{array}{ll}
      (\textit{d} \odot [1,0,1]) \cdot (\textit{n} \odot [1,0,1])
\end{array}  \\
\theta_v^f(d, n)&
=
\begin{array}{ll}
    (\textit{d} \odot [0,1,1]) \cdot (\textit{n}\odot [0,1,1])  \\
\end{array} 
\end{align}
$\textit{d} \in \mathbb{R}^{3}$ stands for the viewing direction vector that starts from the camera positions $c$ and points on a facade point $p$,  the ray direction from the camera to a specific point is computed with the following equation $d=c-p$. The target maps are computed as the dot product of the viewing direction vector $d$ and the facade's surface normal vector $n$. $\theta_h^f$ and $\theta_v^f$ denote the target vectors for the horizontal and vertical axis respectively. $\theta_h^f$ and $\theta_v^f$ target maps differ on the angle that is computed on each occasion, for $\theta_h^f$ targets we consider $x$ and $z$ axis to compute the angle difference on the horizontal axis, while, for vertical maps $\theta_v^f$ we consider $y$ and $z$ axis to output the difference on the vertical axis. To cancel out a specific axis, the Hadamard product $\odot$ is used to isolate either the angle difference on the $x-axis$ for horizontal target vector $\theta_h^f$ or the difference on $y-axis$ for vertical target vector $\theta_v^f$.

This enables a real-time manipulation of textures for an urban 3D scene, more precisely, for each building $b$ we randomly assign a reference image $f$ which serves as texture for this building(cube). In each rendering iteration, our application updates the textures for each building $b$ according to camera location $c$. Given, facade point $p$ their normal $n$, and the camera position $c$ we create the horizontal and vertical view targets $\theta_h$ and $\theta_v$. Then, our model generates the new texture accordingly $t = G(E(f), \theta_{target}^f)$. For more examples check the video in the supplementary materials.

\section{Conclusion and Future Work}
In this paper, we presented FacadeNet, a novel conditional GAN that synthesizes building facade images from different viewpoints given a single input image. By introdcuting the selective editing module, FacadeNet effectively focuses on view-dependent facade features, leading to high-quality synthesized images with fewer artifacts compared to existing methods. Our experimental evaluations demonstrate state-of-the-art performance on standard metrics and appealing qualitative results.

\vspace*{-4mm}
\paragraph*{Future Work}
FacadeNet offers several avenues for future work. First, exploring methods to handle highly complex facades with intricate geometric patterns could enhance the model’s and applicability to a wider range of architectural styles. Moreover, integrating additional context, such as surrounding buildings and natural elements, could improve the visual coherence of the synthesized images in urban environments. FacadeNet could also be extended to support the synthesis of interiors, which would be beneficial for virtual reality applications. Finally, incorporating temporal information for dynamic scene elements (e.g., varying lighting conditions) could make FacadeNet applicable to time-varying view synthesis. 
\vspace*{-5mm}
\paragraph*{Acknowledgements} This project has received funding from
the EU H2020 Research and Innovation Programme and the Republic of Cyprus through the Deputy Ministry
of Research, Innovation and Digital Policy (GA 739578).

{\small
\bibliographystyle{ieee_fullname}
\bibliography{egbib}
}

\vspace{7mm}
\begin{center}
\textbf{\Large -- Supplementary Material --}
\end{center}

\section*{Appendix A: Target viewing vectors} Here we discuss the 
construction process of the target viewing vectors $\mathbf{\theta}_{target}^{gt}$, which are crucial for optimizing FacadeNet. To achieve this, we leverage the pre-processed rectified facades of panoramic images found in the the Large Scale Architectural Asset dataset~\cite{lsaa}. Building facades were already extracted from each panoramic street-view image, during the development of the LSAA dataset. Moreover, for each panorama two rectified 
planes, $\Pi_{rect}^{left}$ and $\Pi_{rect}^{right}$ are constructed, that cover the entire panoramic image. Each rectified 
plane has a predefined width $W_{\Pi}$ and height $H_{\Pi}$ and a horizontal and vertical field of view that span in the
range of $[-75^\circ, +75^\circ]$. Following a standard image rectification process, each extracted facade is mapped to one 
of the two rectified planes, according to its location in the panorama, by maintaining its viewing angles 
$\theta_h^\mathbf{f}$ and $\theta_v^\mathbf{f}$. These angles denote the viewing directions of the center pixel of the 
rectified facade image along the horizontal and 
vertical axes, w.r.t. to a camera that is positioned in the center of the rectified plane, . 

\begin{figure}[!t]
    \centering
    \includegraphics[width=0.49\textwidth]{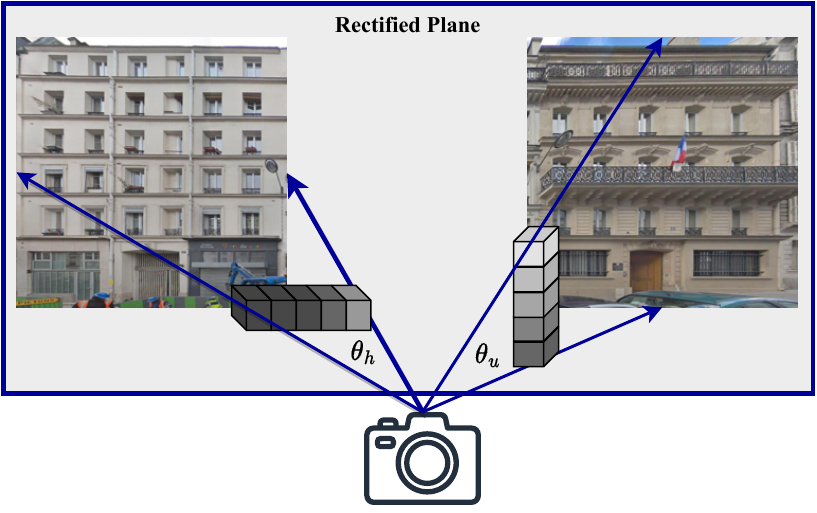}
    \vspace*{-7mm}
    \caption{Facades can occupy various positions on the rectified plane, each position denoting specific horizontal $\theta_h$ and vertical $\theta_v$ view direction targets for the facade image. These view direction targets encompass the location of the facade within the rectified plane, effectively determining the perspective from which it is viewed. By referring to the accompanying figure, it becomes evident that the values of the horizontal and vertical vectors ($h$ and $v$) undergo gradual changes as the points shift across the rectified plane. The spatial location of the rectified plane captures identical view directions for distinct facades}
    \label{fig:va_targets}
    \vspace*{-5mm}
\end{figure}

Utilizing this information, for a rectified plane we construct two horizontal and vertical viewing vectors according to their 
corresponding field of views, normalized in the range $[-1, +1]$, and the spatial dimensionality of each axis, denoted as 
$\mathbf{\theta}_{h}^{\Pi} \in [-1, +1]^{W_{\Pi}}$ and $\mathbf{\theta}_{v}^{\Pi} \in [-1, +1]^{H_{\Pi}}$ respectively. Each 
element of these viewing vectors, captures the horizontal and vertical viewing directions of each pixel in the rectified 
plane. Then, we construct the horizontal target viewing vector $\mathbf{\theta}_{h}^{\mathbf{f}}$ and vertical target 
viewing vector $\mathbf{\theta}_{v}^{\mathbf{f}}$ of the rectified facade, whose dimensionality is equal to the width 
$W_\mathbf{f}$ and $H_\mathbf{f}$ of the facade. Finally, by treating the rectified plane's viewing vectors as lookup 
tables, we find the position of each facade in these, based on the values of $\theta_h^\mathbf{f}$ and 
$\theta_v^\mathbf{f}$, and extract the viewing angles for the facade's target viewing vectors (see Figure \ref{fig:va_targets}). For the synthesis of novel facades, we modify the target viewing vectors of the reference facade by adding a constant negative or positive offset, in order to influence its viewing direction from left to right and top to bottom.
\begin{table*}[tbp]
\centering
\resizebox{\textwidth}{!}{
\begin{tabular}{||c||c c c| c c c c c c ||} 
 \hline
 Method & SEM & features & \# Views & LPIPS-alex$\downarrow$  & LPIPS-vgg$\downarrow$ & FID$_{rec}\downarrow$ & FID$_{novel}\downarrow$ & PSNR$\uparrow$ & SSIM$\uparrow$ \\ [0.5ex] 
 \hline\hline
 $FacadeNet_{base}$ & \textbf{\textendash} & \textbf{\textendash} & 1 & 0.174 & 0.296 & 10.59 & 9.91 & 24.13 & 0.69 \\
 $FacadeNet_{A}$  & \checkmark & semantics & 1 & 0.147 & 0.265 & 9.74  & 9.28 & 24.13 & 0.693 \\
 $FacadeNet_{B}$  & \checkmark & semantics & 2 & 0.143 & 0.262 &  9.66 & 8.89 & 24.25 & 0.698 \\
 $FacadeNet_{C}$  & \checkmark & semantics & 4 & 0.135 & 0.247 &  10.32 & 8.63 & 24.62 & 0.717 \\
 $FacadeNet_{D}$  & \checkmark & semantics & 6 & 0.136 & 0.255 &  9.96 & 9.64 & 24.55 & \textbf{0.718} \\
 $FacadeNet_{E}$  & \checkmark & DINO & 1 & 0.143 & 0.261 & 9.626 & 8.971 & 23.80 & 0.708 \\
 $FacadeNet_{F}$ & \checkmark & DINO & 2 & 0.128 & 0.250 & 9.645 & 8.769 & \textbf{24.77} & 0.712 \\
 $FacadeNet_{G}$  & \checkmark & DINO & 4 & \textbf{0.119} & \textbf{0.240} & \textbf{9.601} & \textbf{8.327} & 23.866 & 0.714 \\
 $FacadeNet_{H}$  & \checkmark & DINO & 6 & 0.145 & 0.265 & 12.39 & 11.453 & 23.91 & 0.708 \\
 \hline
\end{tabular}}
\vspace*{-2mm}
\caption{Our ablation study aims to assess the performance of various design choices we employed in our work.  Specifically, we investigate the impact of \textbf{selective editing mask (SEM)}, the choice of \textbf{features} used as priors for computing the selective mask and the number of novel views per iteration (\textbf{\#Views}). From our findings, we conclude that $SEM$ yields to the most significant improvement in terms of both novel view consistency and image quality. Notably, we observe a substantial enhancement when employing DINO features and learnable weights to combine them, as opposed to manually selected semantic groups as editing masks. Moreover, using more than $1$ view for each iteration improves the results regarding novel inter-view consistency even further.}
\label{table:ablation}
\end{table*}

\begin{figure*}[tbp]
  \centering
  \includegraphics[width=\linewidth]{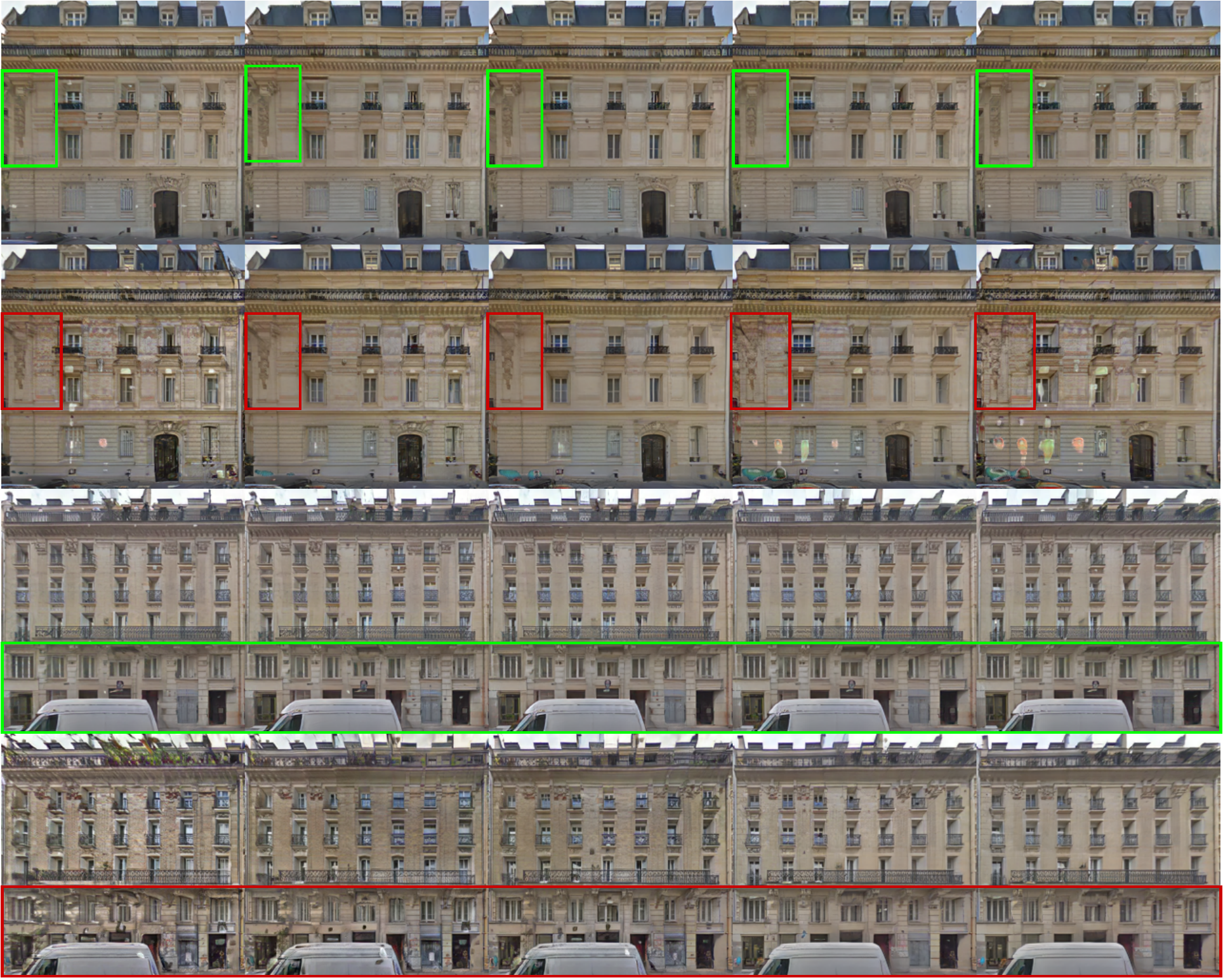}
  \caption{In this figure we display visual comparisons between $FacadeNet_{full}$ and $FacadeNet_{base}$. We observe the emergence of various artifacts when generating examples using $FacadeNet_{base}$ (bottom row of each sample) from different view angles. In contrast, $FacadeNet_{full}'s$ results (top row of each sample) demonstrate a higher level of robustness, effectively preserving the structural details. This distinction is highlighted by the annotated area, where green and red boxes indicate the differences in inter-view consistency.}
  \label{fig:sel_improvments}
\end{figure*}

\begin{figure*}[t]
  \centering
  \includegraphics[width=0.24\linewidth]{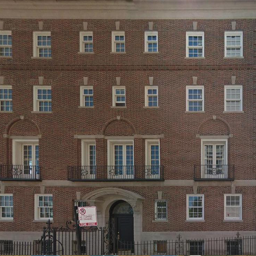}
  \includegraphics[width=0.24\linewidth]{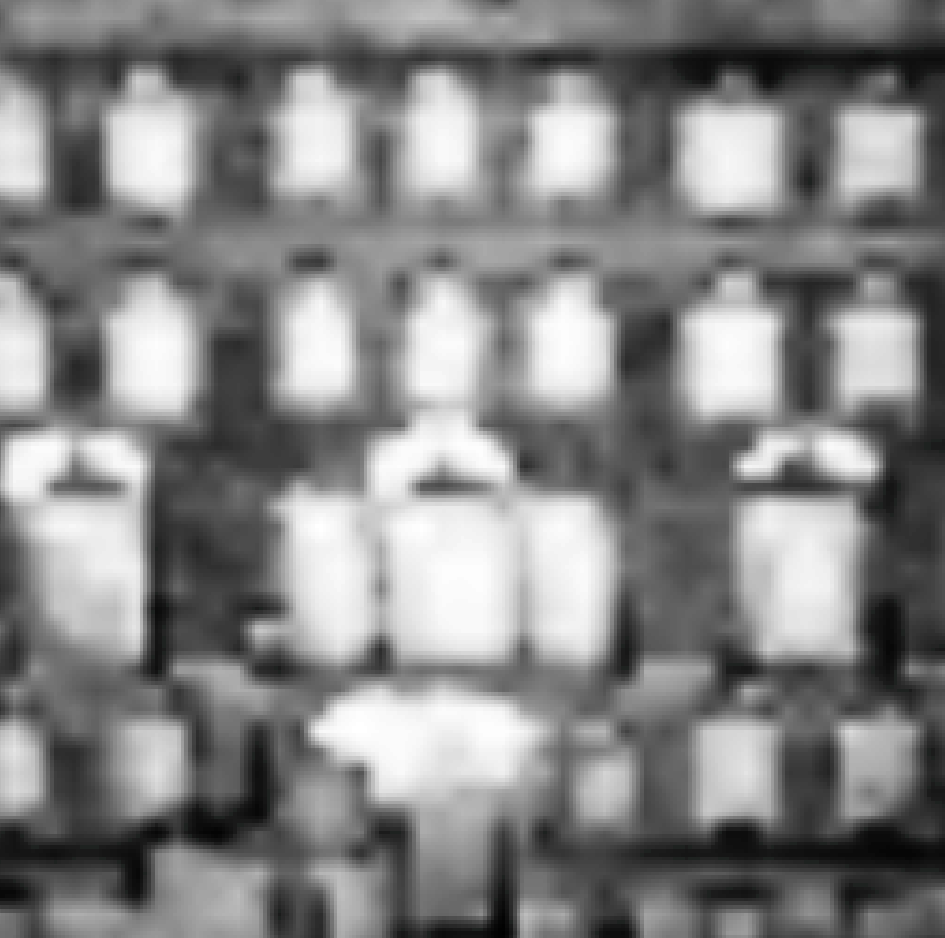}
  \includegraphics[width=0.24\linewidth]{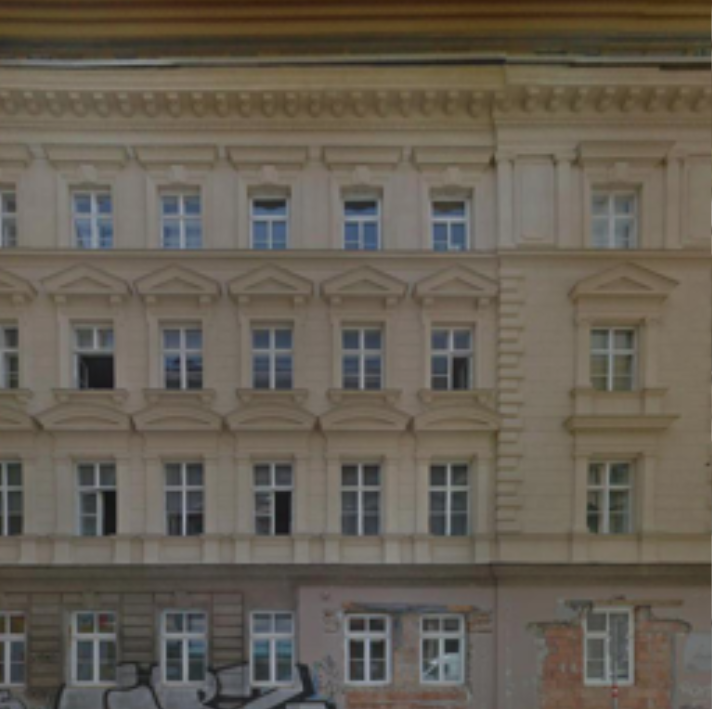}
  \includegraphics[width=0.24\linewidth]{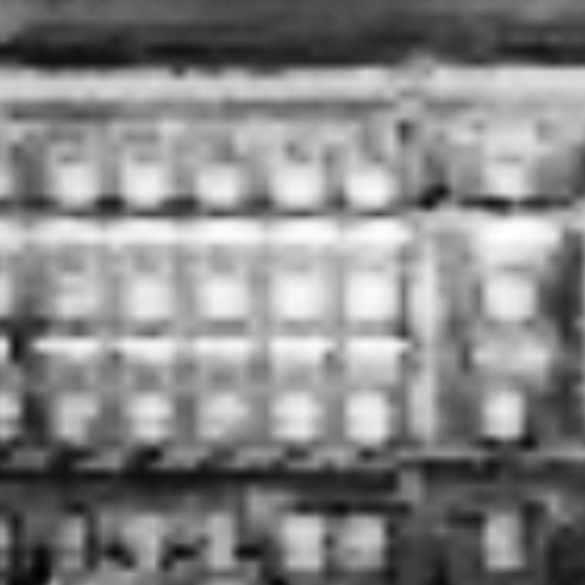} 
  \includegraphics[width=0.24\linewidth]{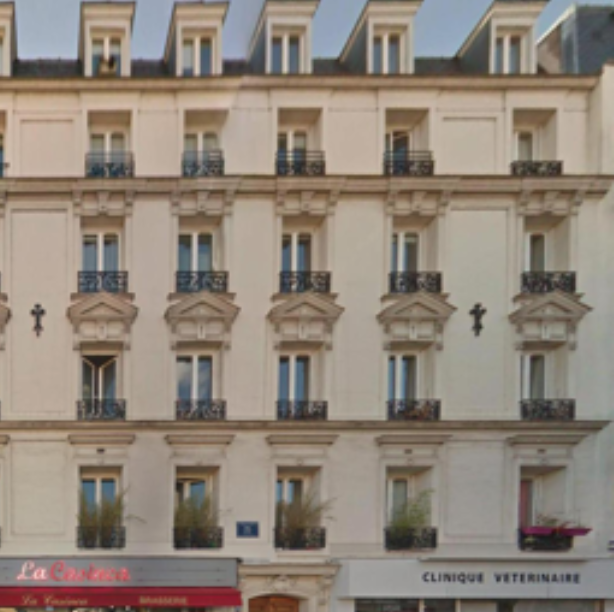}
  \includegraphics[width=0.24\linewidth]{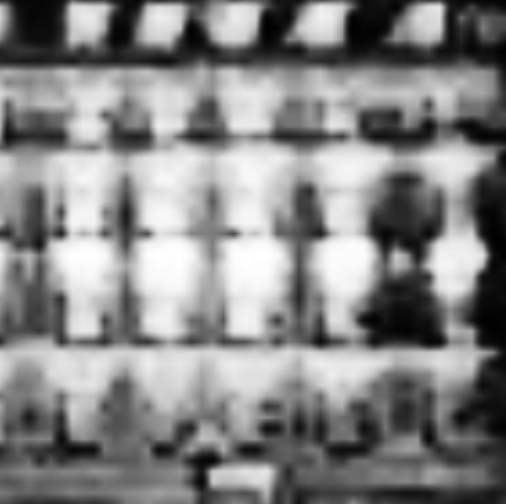}
  \includegraphics[width=0.24\linewidth]{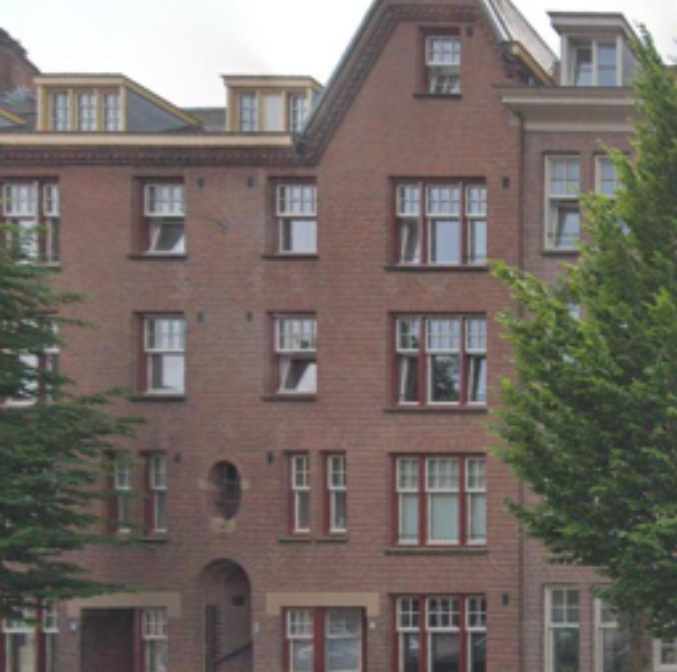}
  \includegraphics[width=0.24\linewidth]{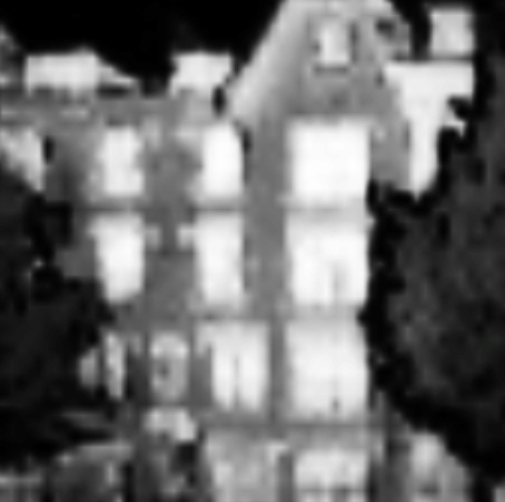}
  \caption{Example of edit masks that were extracted from mask extraction module. }
  \label{fig:supp3_masks}
\end{figure*}

\section*{Appendix B: FacadeNet analysis} 

\paragraph*{Ablation.} We conducted a comprehensive ablation study, presented in Table \ref{table:ablation}, to examine the influence of our design choices on both image synthesis quality and content consistency with respect to reference images. Our model is based on an Encoder-Decoder(Generator) architecture where the reference facade image $f$ is encoded into a latent tensor $z$ that captures the structure and the style of the input image. Subsequently, leveraging the latent matrix $z$, our model generates a coherent representation of the reference facade from various viewpoints by conditioning the generative process using view target vectors.

$FacadeNet_{base}$ serves as our baseline and represents the existing design choice without the selective editing module. Utilizing the selective editing mask (\textbf{SEM}) $FacadeNet_{A}$ achieves higher image quality in novel view synthesis, reducing the metric from $10.59$ to $9.74$. In the case where the \textbf{features} used for the selective editing module is set to \textbf{semantics}, we refer to \textit{fixed masks} that utilize semantic maps and a manually selected group of classes to create a binary mask assigning $1$ to selected classes and $0$ otherwise. Notably, it significantly outperforms the baseline in terms of consistency, with the $LPIPS-alex$ and $LPIPS-vgg$ metrics improving from $0.174$ to $0.147$ and from $0.296$ to $0.265$, respectively. 

$FacadeNet_E$ demonstrates that learnable masks are better suited for the novel view synthesis task. $DINO ViT$\cite{amir2021deep} proves to be a valuable resource, offering meaningful and useful localized features that can effectively be utilized in methodologies similar to ours without the need for supervision. The binary format of fixed masks seems to have a disadvantage in contrast to the continuous representation of masks that are extracted from $DINO ViT$ features \cite{amir2021deep}. Moreover, $DINO ViT$ features provide the freedom to the network to choose the group of features that are required to alter for novel view synthesis in contrast to fixed $semantics$ features. In terms of quality $FacadeNet_{E}$ enhances the $FID_{novel}$ value for novel view synthesis from $9.28$ of $FacadeNet_{A}$ to $8,971$. Additionally, it achieves better scores, reducing the $LPIPS-alex$ metric from $0.147$ to $0.143$ and the $LPIPS-vgg$ metric from $0.265$ to $0.261$. 

Furthermore, it is evident that incorporating multiple views during training brings significant benefits. Intuitively, this approach provides a multi-view consistency, preventing the model from being misled and generating incompatible results between different views. The multi-view version of $FacadeNet$ outperforms their single-view counterparts, regardless, of the other attributes being used in the model. Multi-view training primarily enhances the quality and consistency of novel facade synthesis, while also yielding slight improvements in reconstruction tasks. As illustrated $FacedeNet_{B, C, D}$ clearly outperform $FacedeNet_{A}$ regarding $LPIPS-alex$, $LPIPS-vgg$ and $FID_{novel}$ while the same observation stands for $FacedeNet_{F, G, H}$ in contrast to $FacedeNet{E}$. Among our models, $FacadeNet_{H}$ emerges as the best-performing version. It combines the selective editing module, learnable edit masks and multi-view training, resulting in superior performance compared to other versions of $FacadeNet$. $FacadeNet_{H}$ is referenced as $FacadeNet_{full}$ in the main paper.

\begin{figure}[!t]
    \centering
    \includegraphics[width=0.475\textwidth]{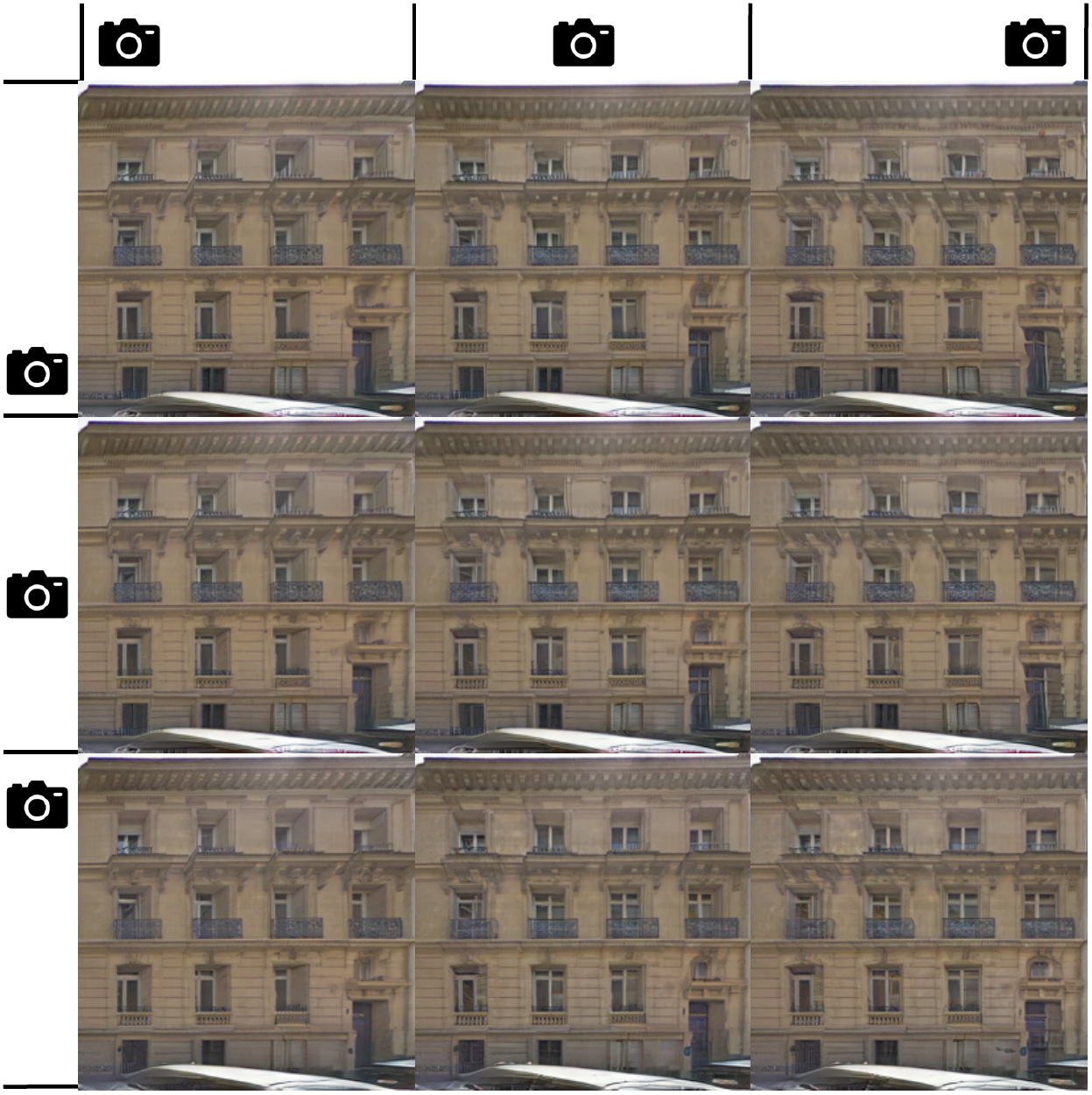}
    \vspace*{-5mm}
    \caption{We present an interpolated representation of camera movement in both horizontal and vertical directions. The spatial code $z$ remains fixed, while the output dynamically adjusts according to the camera's motion, which modifies the conditional information of the view direction target, $\theta_{target}$. In the accompanying figure, we demonstrate the camera's movement from left to right (top row) along the horizontal axis and from top to bottom (left column) along the vertical axis. Furthermore, we showcase the disentangled controllability achieved on both axes.}
    \vspace*{-5mm}
    \label{fig:facade_interpolation}
\end{figure}

\paragraph*{Selective editing improvements.} To visually assess the effectiveness of our selective editing module, we conducted a comparison between our $FacadeNet_{full}$ and our base model, $FacadeNet_{base}$, in order to validate the improvements in consistency. In Figure \ref{fig:sel_improvments}, we present an interpolation of view angles for two facades, emphasizing the enhancements we have achieved.

While both models generate high-quality and believable center images, it is evident that the model trained with the selective editing module exhibits significantly superior consistency. The improvements in maintaining coherence and smooth transitions between the generated views are remarkable when compared to our base model.

In Figure \ref{fig:sel_improvments}, we present the results of $FacadeNet_{full}$ and $FacadeNet_{base}$ in pairs of rows. The top row in each sample corresponds to the outputs generated by $FacadeNet_{full}$, while the rows below depict the results from $FacadeNet_{base}$. Upon careful observation, it becomes apparent that $FacadeNet_{base}$ introduces various artifacts between different view angles. In contrast, $FacadeNet_{full}$ demonstrates a higher level of robustness and maintains the integrity of facades' detailed areas (see highlighted area in the green and red boxes in Figure 7).

In the second sample (rows 3 \& 4), $FacadeNet_{full}$ exhibits the ability to discern insignificant features, such as the car that is present in the image, and preserves them consistently across different views. However, $FacadeNet_{base}$ fails to maintain such details, resulting in distorted and peculiar outcomes (see last row red box in figure \ref{fig:sel_improvments}).

\begin{figure*}[h]
  \centering
  \includegraphics[width=\linewidth]{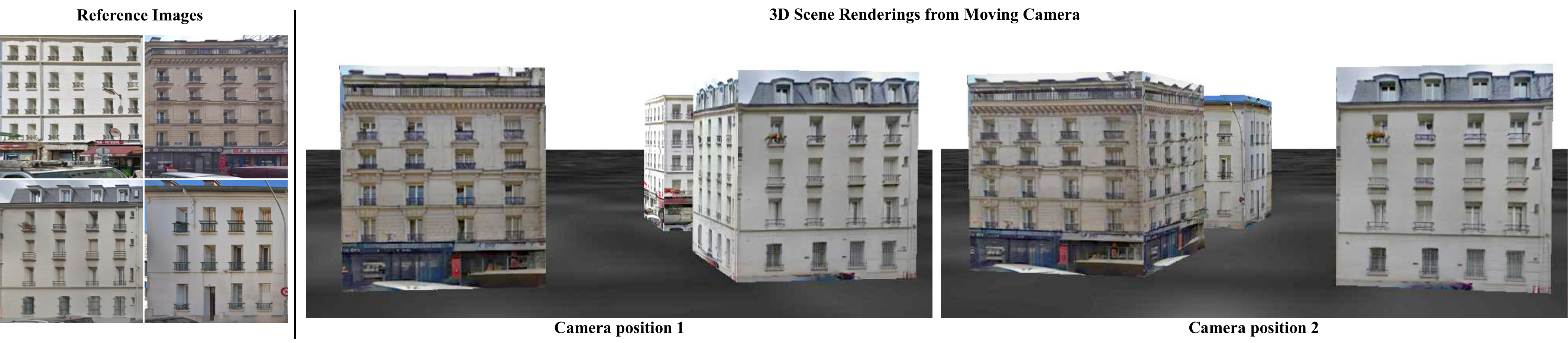}
  \caption{Urban Scenes Renders from the in  Real-Time interactive textures application}
  \label{fig:inter}
\end{figure*}

\begin{figure*}[t]
  \centering
  \includegraphics[width=0.49\linewidth]{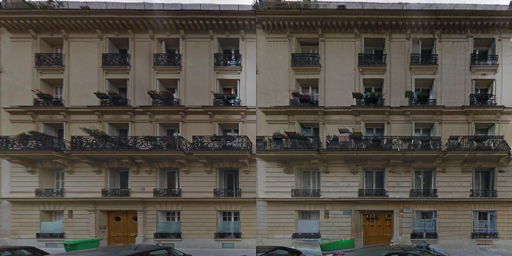}
  \includegraphics[width=0.49\linewidth]{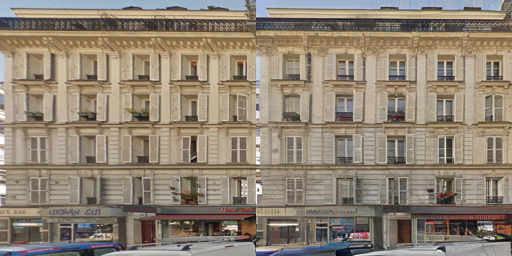}
  \includegraphics[width=0.49\linewidth]{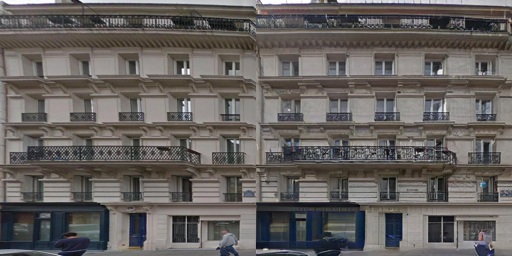}
  \includegraphics[width=0.49\linewidth]{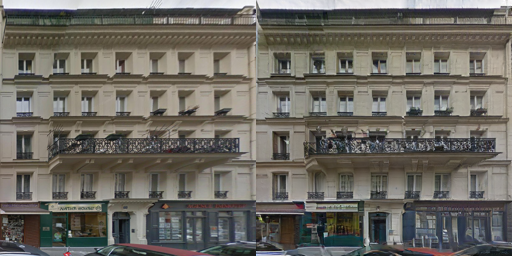}
    \vspace*{-1mm}
  \caption{We display examples of problematic facade image improvement. We display pairs of the reference images (left) and the $\theta_0$(center image) reconstruction images(right). We observe that our model can rotate the facade to a better view orientation in contrast to the reference while at the same time, it achieves a high similarity of style and structure.}
  \label{fig:facade_undistortion}
\end{figure*}

\begin{figure*}[h]
  \centering
  \includegraphics[width=\linewidth]{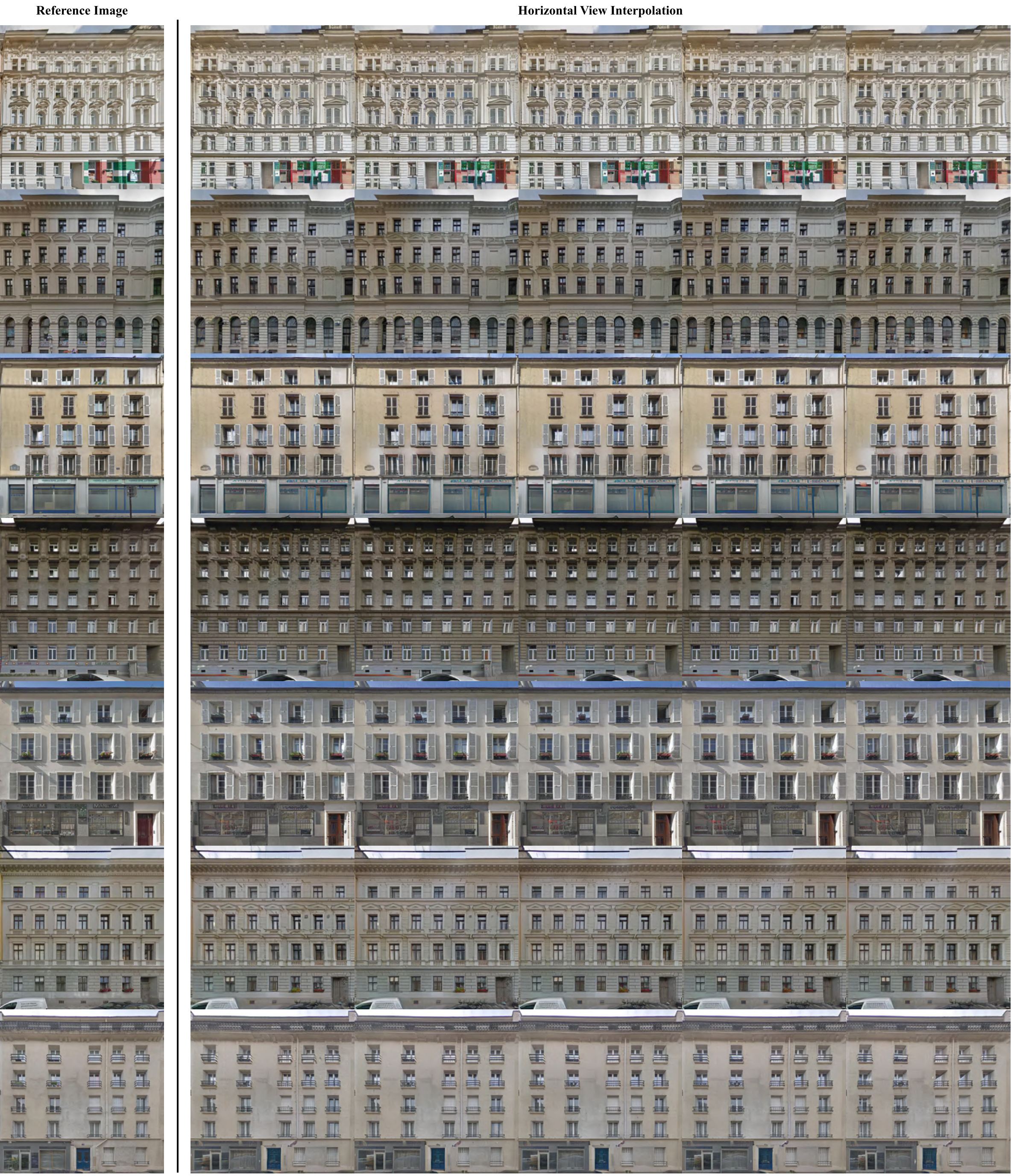}
  \caption{Examples of 5-step image interpolation on the horizontal axis. Given the reference images (left column), we can reconstruct the novel view from different angles as it is illustrated in the images of columns 2-6.}
  \label{fig:supp1_results}
\end{figure*}

\begin{figure*}[h]
  \centering
  \includegraphics[width=\linewidth]{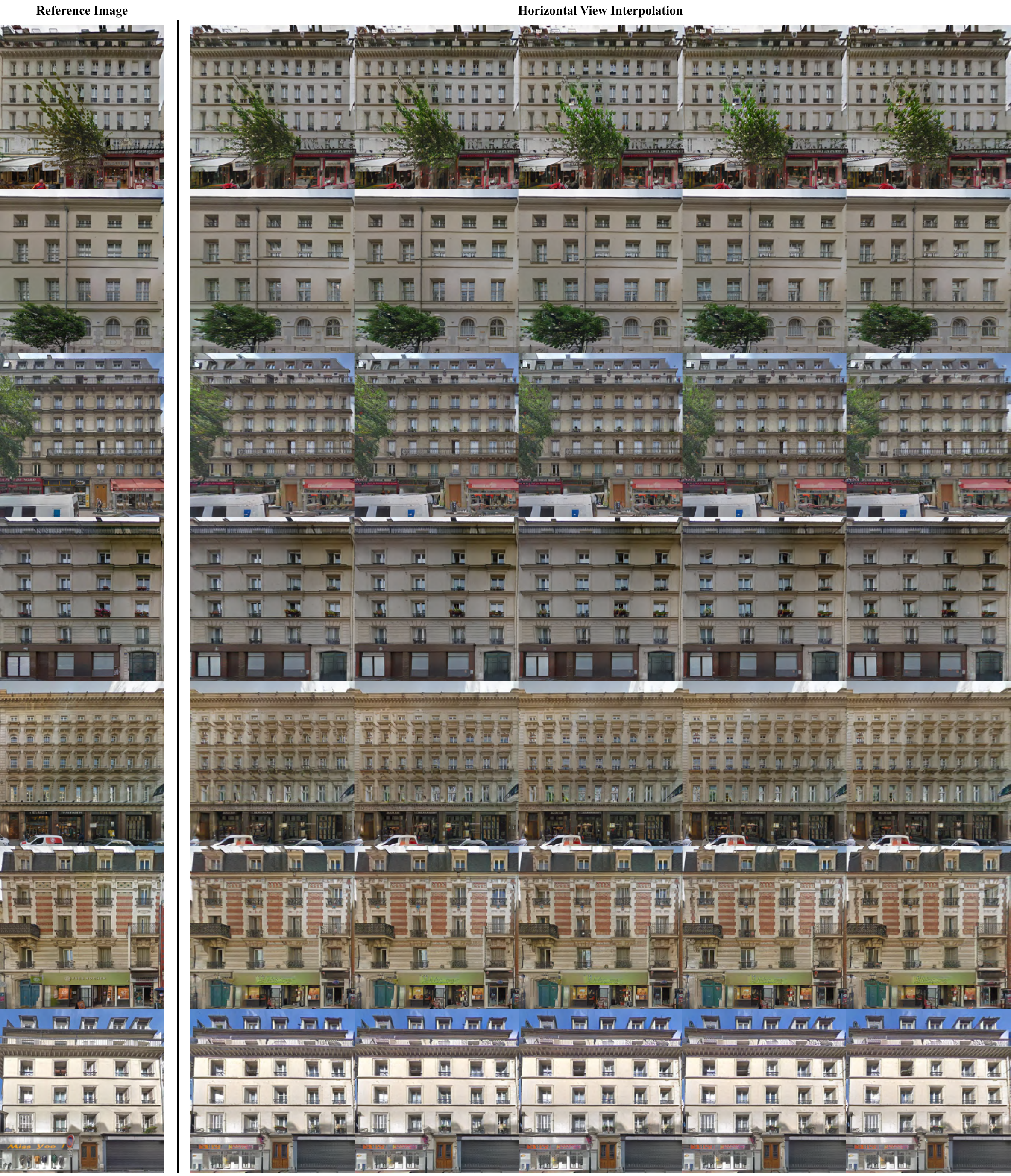}
  \caption{Examples of 5-step image interpolation on the horizontal axis. Given the reference images (left column), we can reconstruct the novel view from different angles as it is illustrated in the images of columns 2-6.}
  \label{fig:supp2_results}
\end{figure*}

\section*{Appendix C: Facade view interpolation}\label{sec:multi-view}

Here, we demonstrate the impact of our horizontal vector $\theta_h$ and vertical vector $\theta_v$ on specific input facade images, showcasing their influence. By utilizing our encoder, we obtain a latent matrix $z$ based on a reference facade image $f_{ref}$. Subsequently, our generator employs the same latent matrix $z$ along with different combinations of horizontal $\theta_h$ and vertical $\theta_v$ targets to generate samples.

Figure \ref{fig:facade_interpolation} illustrates the camera movement on the horizontal axis, moving from right to left (represented by the camera's motion in the top row). Similarly, on the vertical axis, the camera moves from bottom to top (represented by the camera's motion in the left column). This visual representation exemplifies the disentangled controllability of the view angle vectors for both axes. As evident from the results, the generated content precisely aligns with the target vectors.

Figures \ref{fig:supp1_results} and \ref{fig:supp2_results} present additional results that serve to exemplify the effectiveness of our approach. These visual examples showcase our model's ability to successfully handle a diverse range of structures and architectural styles. Notably, our method maintains the overall style coherence while preserving the distinctive style of individual windows throughout the interpolation process. Moreover, we observe the robustness of our model in challenging scenarios where the image contains noise or when facade details are partially occluded by trees. These results highlight the adaptability and reliability of our approach in real-world situations.

\section*{Appendix D: FacadeNet applications results}

\paragraph*{Problematic rectified facade improvement.} Figure \ref{fig:facade_undistortion} showcases pairs of facade images for visual comparison. The top row displays the problematic rectified facade images $f_{ref}$, while the bottom row exhibits the 0-view improved generated facade images $f_{novel} = G(E(z_f, \theta_0))$. We observe that our generative approach successfully reconstructs an identical appearance to the reference facade images $f_{ref}$ while maintaining structure and style. Additionally, the approach effectively translates the components  of the facades to align with a 0-angle viewpoint.  This transformation results in fewer problematic areas in the generated facade images.

\paragraph*{Real-time textures for urban scenes.}
In Figure \ref{fig:inter} we render examples of our approach from $2$ different view angles. This example contains $4$ buildings whose textures are changing simultaneously but differently based on their location in the 3D world and the position of the camera. We illustrate that our application can create multiple plausible results in real-time

\section*{Appendix E: Additional visualizations}

\paragraph*{Edit mask examples.} Figure \ref{fig:supp3_masks} provides a visual representation of the edit mask obtained through our selective editing module. This module leverages information from DINO ViT to extract plausible edit masks by employing learnable weights that blend the input features into a 1-channel edit mask. The purpose of this edit mask is to guide the network in manipulating the reference image in a targeted manner, thereby enhancing the consistency of novel views across different view angle targets.

A notable observation in Figure \ref{fig:supp3_masks} is the consistent pattern exhibited by the selective editing module. It assigns high values to semantic areas such as windows, doors, and balconies, indicating their significance in the editing process. Moderate values are assigned to various facade details, particularly those found on the ground floor. Areas with plain walls or minimal details receive low importance values, while the sky and trees receive the lowest importance values in the context of the novel view synthesis task.

The underlying rationale behind the use of masks is to group areas in the reference image that are crucial for our task and focus on modifying them while leaving the remaining areas intact. This approach allows us to selectively and effectively alter specific regions of the image to achieve our desired outcomes.

\begin{figure*}[tbp]
  \centering
  \includegraphics[width=\linewidth, height=9 cm]{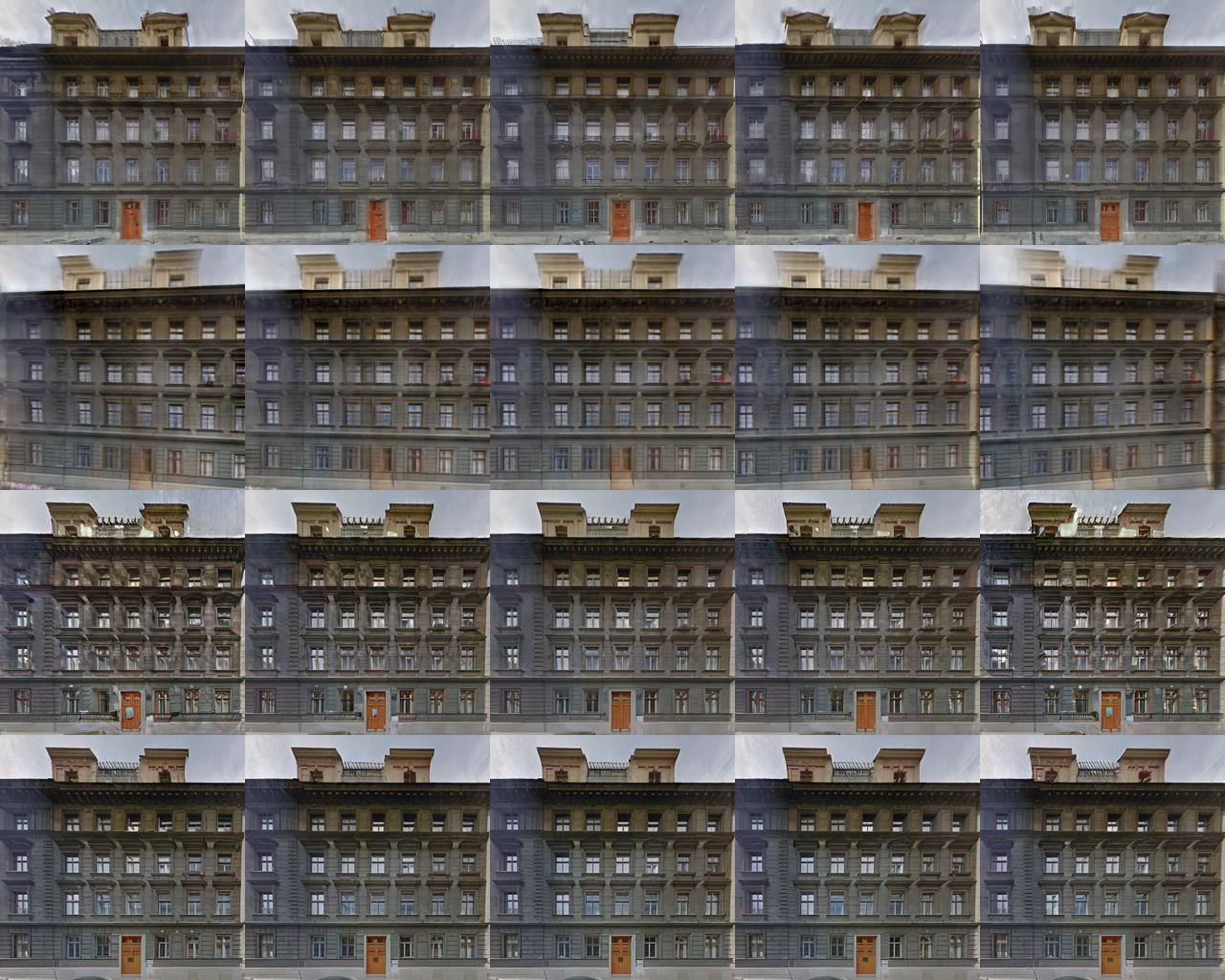}
  \caption{Novel view interpolation comparison between textit{Palette}~\cite{saharia2021palette} ($1^{st}$ row), \textit{3DGP}~\cite{skorokhodov20233d} ($2^{nd}$ row), \textit{swapping-AE}~\cite{park2020swapping} ($3^{rd}$ row) and $FacadeNet_{full}$ ($4^{th}$ row)}
  \label{fig:interpolation_1}
\end{figure*}

\begin{figure*}[tbp]
  \centering
  \includegraphics[width=\linewidth, height=9 cm]{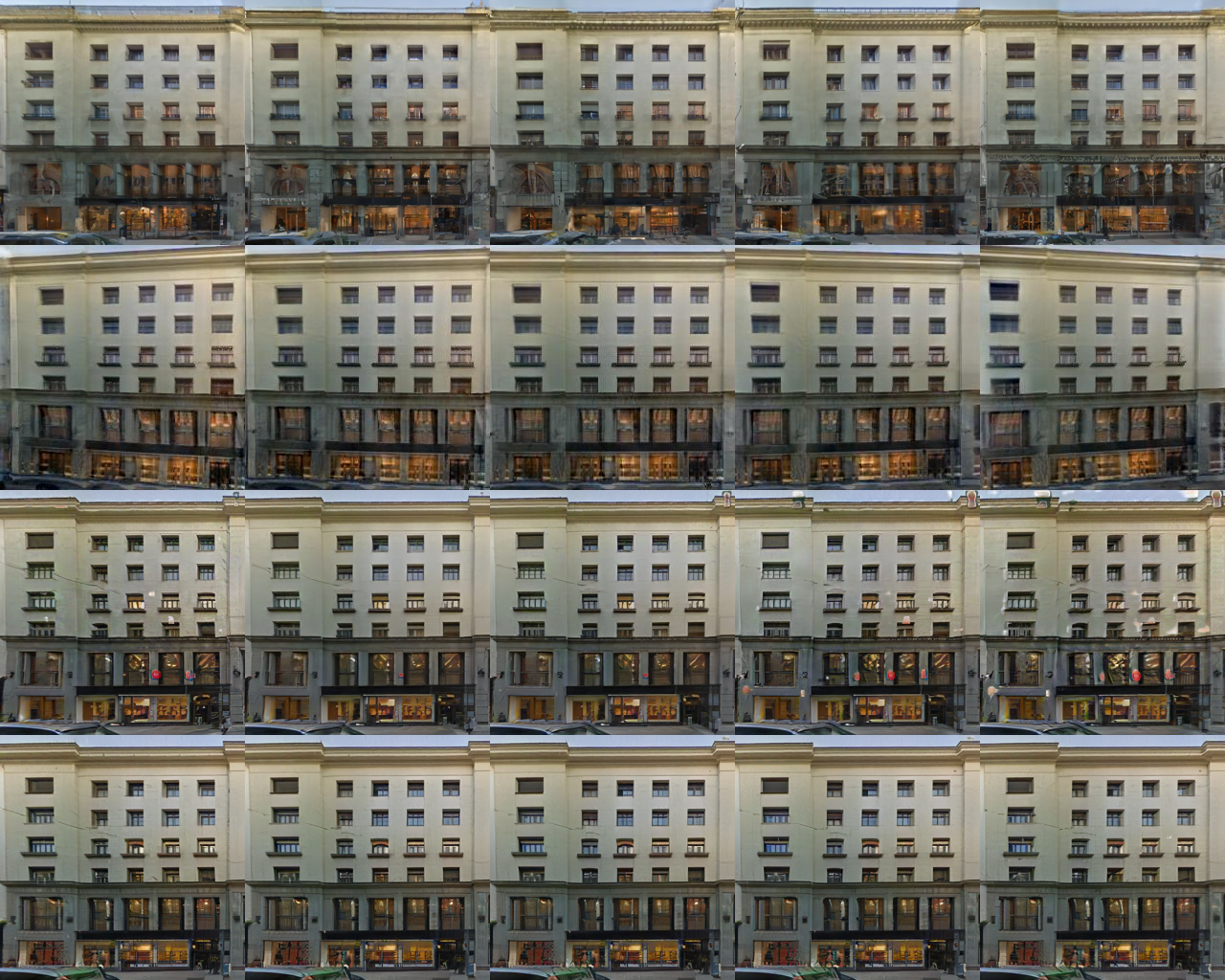}
  \caption{Novel view interpolation comparison between textit{Palette}~\cite{saharia2021palette} ($1^{st}$ row), \textit{3DGP}~\cite{skorokhodov20233d} ($2^{nd}$ row), \textit{swapping-AE}~\cite{park2020swapping} ($3^{rd}$ row) and $FacadeNet_{full}$ ($4^{th}$ row)}
  \label{fig:interpolation_2}
\end{figure*}

\begin{figure*}[tbp]
  \centering
  \includegraphics[width=\linewidth, height=9 cm]{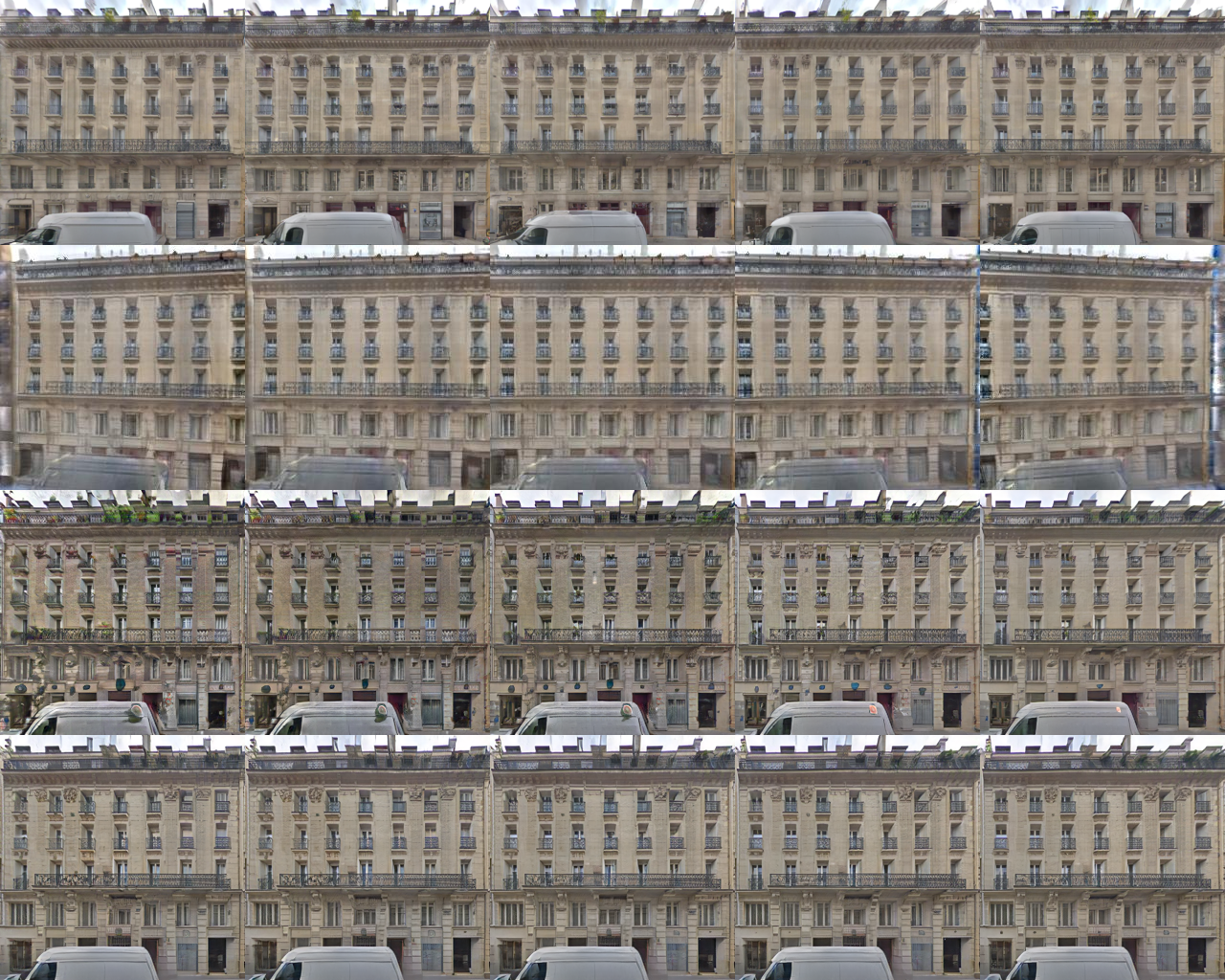}
  \caption{Novel view interpolation comparison between textit{Palette}~\cite{saharia2021palette} ($1^{st}$ row), \textit{3DGP}~\cite{skorokhodov20233d} ($2^{nd}$ row), \textit{swapping-AE}~\cite{park2020swapping} ($3^{rd}$ row) and $FacadeNet_{full}$ ($4^{th}$ row)}
  \label{fig:interpolation_3}
\end{figure*}

\begin{figure*}[tbp]
  \centering
  \includegraphics[width=\linewidth, height=9 cm]{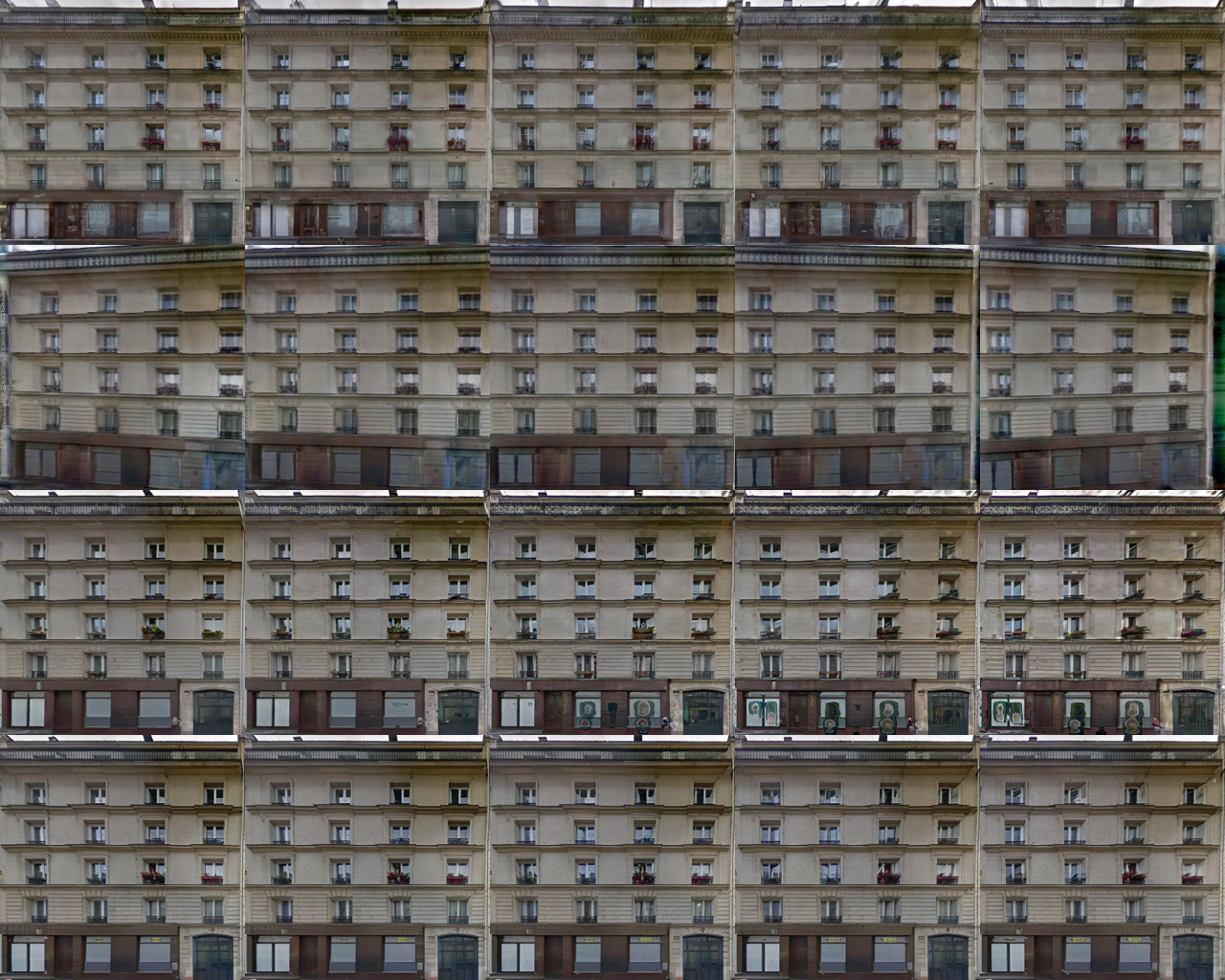}
  \caption{Novel view interpolation comparison between textit{Palette}~\cite{saharia2021palette} ($1^{st}$ row), \textit{3DGP}~\cite{skorokhodov20233d} ($2^{nd}$ row), \textit{swapping-AE}~\cite{park2020swapping} ($3^{rd}$ row) and $FacadeNet_{full}$ ($4^{th}$ row)}
  \label{fig:interpolation_4}
\end{figure*}

\paragraph*{Qualitative results.} Figures \ref{fig:interpolation_1}, \ref{fig:interpolation_2}, \ref{fig:interpolation_3}, \ref{fig:interpolation_4} presents qualitative comparisons between \textit{Palette}\cite{saharia2021palette} ($1^{st}$ row), \textit{3DGP}\cite{skorokhodov20233d} ($2^{nd}$ row), \textit{swapping-AE}\cite{park2020swapping} ($3^{rd}$ row) and $FacadeNet_{full}$ ($4^{th}$ row). \textit{Palette} and \textit{3DGP} are unable to generate fine details as the generation is combined with novel view synthesis. Notably, artifacts become apparent in the output generated by the $swapping-AE$ model across varying viewing angles. In contrast, $FacadeNet_{full}$'s results  demonstrate a higher level of robustness, effectively preserving the structural details. More results are displayed in the supplementary.

\end{document}